\documentclass{article}

\usepackage[preprint]{neurips_2026}
\PassOptionsToPackage{numbers,compress}{natbib}
\usepackage[utf8]{inputenc}
\usepackage[T1]{fontenc}
\usepackage{hyperref}
\usepackage{url}
\usepackage{booktabs}
\usepackage{amsfonts}
\usepackage{nicefrac}
\usepackage{microtype}
\usepackage{xcolor}
\usepackage{graphicx}
\usepackage{amsmath,amssymb}
\usepackage{multirow}
\usepackage{array}
\usepackage{enumitem}
\usepackage{float}
\usepackage{hyperref}       

\usepackage{subcaption}
\usepackage{placeins}
\usepackage{threeparttable}
\usepackage{tabularx}
\usepackage{algorithm}
\usepackage{algorithmic}

\title{FlexMS: A Unified Public Benchmark for Molecule Tandem Mass Spectrum Prediction} 

\author{
  Yunhua Zhong$^{1,3}$\thanks{This work was done during an internship at HKUST-GZ.},
  Yixuan Tang$^{1}$,
  Yifan Li$^{1}$,
  Pan Liu$^{1}$, \\
  \textbf{Zhiwen Yang}$^{1,4}$
  \textbf{Zanyi Wang}$^{6}$,
  \textbf{Jie Yang}$^{5}$,
  \textbf{Jun Xia}$^{1,2}$\thanks{Corresponding author.} \\
  $^{1}$The Hong Kong University of Science and Technology (Guangzhou).\\
  $^{2}$The Hong Kong University of Science and Technology.\\ 
  $^{3}$The University of Hong Kong. $^{4}$Yangzhou University. $^{5}$Fudan University. \\
  $^{6}$ University of California, San Diego
  \vspace{0.1cm} \\ 
  Corresponding author(s). E-mail(s): \href{mailto:junxia@hkust-gz.edu.cn}{junxia@hkust-gz.edu.cn}; \\
  Contributing authors: \href{mailto:yunhuazhong@connect.hku.hk}{yunhuazhong@connect.hku.hk};\href{mailto:tangyixuan@stu2022.jnu.edu.cn}{tangyixuan@stu2022.jnu.edu.cn}\\
  \href{mailto:yli994@connect.hkust-gz.edu.cn}{yli994@connect.hkust-gz.edu.cn}; \href{mailto:panliu@hkust-gz.edu.cn}{panliu@hkust-gz.edu.cn}; \\
  \href{mailto:zhiwenyang2004@gmail.com}{zhiwenyang2004@gmail.com}; \href{mailto:yangjie23@m.fudan.edu.cn}{yangjie23@m.fudan.edu.cn}; 
}

\begin{document}

\maketitle

\begin{abstract}
Tandem mass spectrometry (MS/MS) is central to small molecule identification, but current deep learning systems for spectrum prediction still remain difficult to evaluate and deploy in practice. While novel architectures constantly claim state-of-the-art performance, inconsistent metadata conditioning and entangled preprocessing pipelines hinder fair architectural comparisons. Besides, existing evaluations are often restricted to curated datasets, failing to capture the heterogeneity and cross-domain shifts of real-world metabolomics. Furthermore, current benchmarks lack difficulty-aware diagnostics and leave blind to how models behave under specific compute or data constraints. To address this, we present FlexMS, a modular public-data benchmark framework that standardizes MS/MS prediction across public resources while keeping molecular encoders, metadata conditioning, predictor heads, and downstream retrieval under one protocol. FlexMS establishes a fair evaluation playground which significantly lowers the barrier for integrating new predictive tools. Rather than solely optimizing for average scores, FlexMS augments aggregate accuracy with difficulty-aware diagnostics, providing actionable guidance on model selection across different compute constraints, data scales, and downstream retrieval objectives. Ultimately, FlexMS provides the community with a reproducible standard to identify which algorithmic conclusions are stable and which operating points are most viable in practice.
\end{abstract}


\section{Introduction}
\label{sec:intro}

Tandem mass spectrometry (MS/MS) is central to molecule identification, drug discovery, and metabolism analysis because fragmentation patterns can provide rich clues about molecular structure \citep{aebersold2016mass,de2007mass}. In practical analytical workflows, interpreting an unannotated spectrum frequently requires evaluating multiple plausible structural hypotheses, such as proposed drug metabolites, reaction byproducts or novel natural product derivatives. Computational spectrum prediction and retrieval serve as crucial diagnostic tools in this context because researchers can perform \textit{in silico} retrieval to rank candidates against the experimental signal \citep{duhrkop2019sirius, kind2018identification, wishart2018hmdb}. Formally, the computational task of \textit{in silico} MS/MS spectrum prediction maps a molecular structure (e.g., SMILES or 2D graph) and essential experimental metadata (e.g., collision energy, precursor ion and charge) to a simulated fragmentation spectrum. To make the output space tractable for neural networks, the continuous $m/z$ (mass-to-charge ratio) axis is typically discretized into fixed bins, which transforms the task into a high-dimensional regression problem. The model predicts the relative intensity of fragment ions within each bin.

To solve this high-dimensional regression problem effectively, the field has increasingly shifted away from traditional rule-based empirical methods toward deep learning architectures\citep{wang2021cfmid4, ruttkies2016metfrag}. Because these DL models excel at learning complex fragmentation heuristics directly from large-scale mass spectral databases, the model space for spectrum prediction has expanded dramatically. Recent methods now span descriptor-based MLPs, sequence models over SMILES, graph neural networks and pretrained chemical foundation models \citep{chithrananda2020chemberta, goldman2024generating, hirohara2018convolutional, wei2019rapid, young2024tandem}. In other words, the field no longer lacks modeling ideas.

\begin{figure}[t]
    \centering
    \includegraphics[width=14cm]{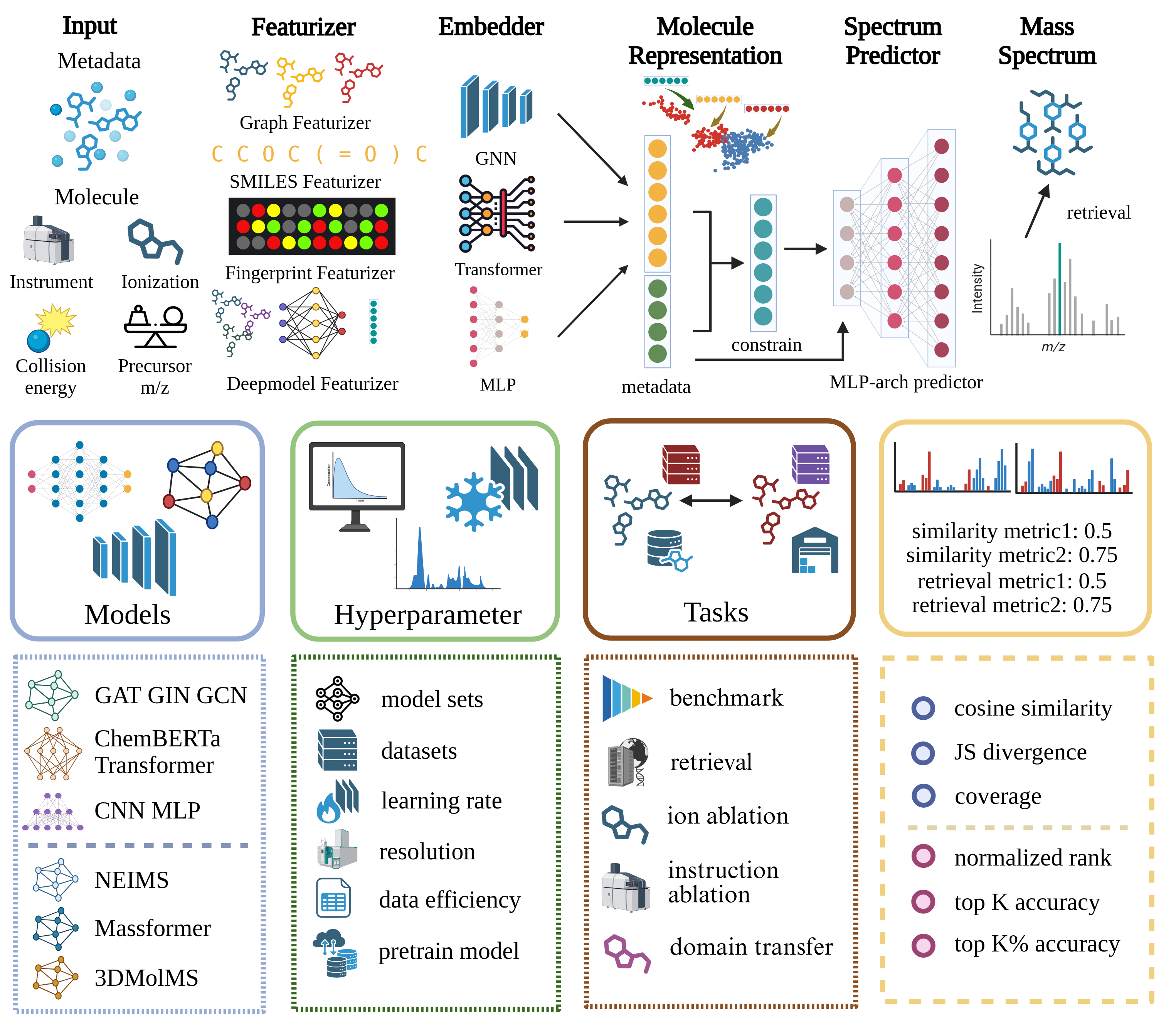}
    \caption{Main components of FlexMS. FlexMS framework takes molecules and associated metadata as inputs, employs various featurizers and embedders to generate molecular representations, and utilizes different multi-layer perceptron (MLP) architectures for spectrum prediction at specified bin resolutions. We assess the performance of various models, investigate the impact of different hyperparameters, and evaluate outcomes across diverse task scenarios. Comprehensive metrics are employed to quantify performance.}
    \label{fig:FlexMS-main}
\end{figure}

This lack of a shared protocol is one of the main bottlenecks in modern spectrum prediction. A typical pipeline combines a molecular representation choice, a metadata-conditioning scheme, a molecule-to-spectrum prediction head, a target representation rule, and a downstream optimization and evaluation objective\citep{young2024tandem}. New methods are often assembled \textit{de novo} rather than within a standardized framework, which makes benchmarking fragile\citep{murphy2023graffms}. In practice, model family, metadata assumptions, split semantics, and evaluation targets often vary together, making reported gains difficult to attribute to model quality alone.

Current evaluation practice exhibits three concrete limitations. First, many studies provide incomplete protocol coverage, reporting only few datasets, a single task and a limited set of prediction metrics, even though different evaluation settings can support very different conclusions about generalization\citep{murphy2023graffms}. Second, comparisons are often entangled as reported gains may be confounded by preprocessing pipelines, metadata availability, or inaccessible assets, and some reported results depend on private commercial data or expensive substructure-annotation pipelines that limit fair comparison and efficient reuse \citep{goldman2024generating,young2024tandem,goldman2023prefix,nist20,ridder2014magma,hong20233dmolms}. Finally, many models provide limited or hard practical guidance, as they typically optimize for the best average score but offer less insight into which model family remains preferable when data are limited, compute is constrained, or the downstream objective shifts from spectrum reconstruction to molecule retrieval.

Similar issues in chemical machine learning have motivated benchmark-centered resources where progress depends not only on stronger models, but also on stronger evaluation protocols \citep{liu2024flexmol,wu2018moleculenet}. While recent works like MassSpecGym \citep{bushuiev2024massspecgym} have made significant strides in standardizing MS/MS datasets and defining prediction tasks, FlexMS complements these efforts by providing a more modular and diagnostic-oriented infrastructure. Specifically, while MassSpecGym focuses on curated tasks and splits, FlexMS adds modular encoder-predictor construction, multi-dataset diagnostics, compute-aware analysis, and retrieval evaluation. Since researchers often implement deeply coupled, custom scripts to conduct experiments, FlexMS addresses this critical gap by treating MS/MS prediction not merely as a leaderboard competition, but as a flexible and modular infrastructure problem over shared public resources. Rather than narrowly asking which model achieves the highest nominal score, FlexMS provides an aligned evaluation playground that yields actionable insights. By systematically mapping these aligned metrics to specific split difficulties, data regimes and compute budgets, FlexMS elevates difficulty-aware evaluation to serve as both a rigorous tool and a practical guide, helping practitioners to decide which architecture to deploy.



\textbf{Contributions}
\begin{itemize}
  \item We introduce FlexMS as a modular public-data benchmark framework for MS/MS prediction, designed so that new datasets, models and evaluations can be inserted into one reproducible comparison space.
  \item We define a unified cross-dataset protocol, including official or scaffold-aware splits, metadata-aware conditioning, and shared similarity metrics.
  \item We contribute a difficulty-aware evaluation view for tandem MS prediction by coupling aggregate benchmark scores with molecular-similarity and spectral-entropy diagnostics, so that changes in model ranking can be interpreted rather than merely observed.
  \item We distill practical, constraint-aware insights to help practitioners identify optimal model choices and operating points based on specific data regimes, computational budgets and evaluating settings.
\end{itemize}

\section{Related Work}
\label{sec:related}

\textbf{Spectrum prediction methods.}
Deep learning has turned MS/MS prediction into a supervised mapping from molecular structure to spectral outputs. Existing approaches now span descriptor-based predictors such as NEIMS \citep{wei2019rapid}, graph-transformer systems such as MassFormer \citep{young2024tandem}, geometry-aware models such as 3DMolMS \citep{hong20233dmolms}, and autoregressive fragmentation predictors \citep{goldman2023prefix,goldman2024generating}. At the representation level, researchers also draw from a broad encoder space including classical graph neural networks, attentive molecular encoders, and pretrained chemical foundation models \citep{chithrananda2020chemberta,kipf2016semi,rong2020self,velivckovic2017graph,xia2023mole,xiong2019pushing,xu2018powerful}. This diversity is scientifically useful but makes fair comparison difficult: when datasets, split semantics, metadata assumptions, and preprocessing pipelines differ across papers, improvements in model design do not translate into stable conclusions about model choice.

\textbf{Public spectral libraries.}
GNPS \citep{wang2016sharing}, MassBank \citep{horai2010massbank}, and NIST \citep{nist20} are the most widely used MS/MS resources, but they differ substantially in scale, curation quality and licensing. GNPS offers the largest crowd-sourced collection with substantial heterogeneity with official preprocess version; MassBank provides cleaner but source-diverse reference spectra at smaller scale; NIST remains difficult to use in fully reproducible public benchmarking because of commercial restrictions. As a result, many prediction papers rely on overlapping raw resources but adopt different filtering rules, metadata handling and train--test splits, so even nominally similar evaluations can correspond to materially different benchmark conditions.

\textbf{Benchmarks and modular frameworks.}
Benchmark-centered resources in cheminformatics such as MoleculeNet \citep{wu2018moleculenet}, Therapeutics Data Commons \citep{huang2021therapeutics}, and ProteinGym \citep{notin2023proteingym} have shown that community progress depends not only on stronger models, but also on shared datasets, standardized splits and protocols that make conclusions comparable across studies. FlexMol \citep{liu2024flexmol} further showed that modular frameworks enabling systematic encoder--predictor combinations can make comparisons broader and fairer. Within MS/MS, MassSpecGym \citep{bushuiev2024massspecgym} unifies multiple annotation challenges around one curated resource and a demanding split design. These cross-dataset, protocol-centered works emphasize recommendations and difficulty-aware diagnostics rather than only reporting leaderboard.


\section{FlexMS Framework}
\label{sec:methods}

\subsection{Mass Spectrum Prediction}
Inputs for mass spectrum prediction consist of molecular entities and their associated experimental metadata \cite{bushuiev2024massspecgym}. Molecules are typically represented as SMILES strings or 2D/3D molecular graphs. The experimental metadata includes conditions such as collision energy (CE), instrument type, and precursor ion mode. The primary objective is to develop a mapping function \( f: (X, M) \rightarrow Y \) that takes a molecular representation \( X \) and metadata embedding \( M \) as inputs, and predicts the interaction fragmentation pattern \( Y \). In our binned prediction approach, the continuous mass-to-charge (\(m/z\)) domain is discretized into fixed-width bins (e.g., 1 Da), transforming the task into a multi-label regression problem where the model outputs intensity values for each predefined bin.

We provide a detailed ablation on how varying this bin width from 1 Da to 10 Da non-linearly affects different model architectures in \textit{Appendix}~\ref{app:extended_conclusions}.

\subsection{Framework}
FlexMS provides a modular and user-friendly API for the dynamic construction of mass spectrum prediction models. The workflow consists of three primary stages: First, users select a specific task and load the corresponding dataset with designated split strategies (random or scaffold). Second, users customize their models by declaring and combining FlexMS components, which strictly decouple the \texttt{Embedder} for molecular representation, the \texttt{Predictor} for spectrum generation along with a \texttt{Metadata Encoder}. Third, FlexMS automatically handles data featurization, constructs the computational graph, and manages training and multi-metric evaluations.

\subsubsection{Components}

\textbf{Featurizer} \quad This module handles the transformation of raw chemical inputs into specific mathematical formats required by downstream models. It generates diverse featurizations, including fixed-length physicochemical descriptors and binary fingerprints (e.g., Morgan, MACCS), tokenized sequences for language models (e.g., via byte-pair encoding), and topological graph objects with detailed atomic node and bond edge features for message-passing networks.

\textbf{Metadata Encoder} \quad The Metadata Encoder standardizes continuous experimental variables such as Absolute Collision Energy (ACE) or Normalized Collision Energy (NCE), Discretized precursor $m/z$, and one-hot encodes categorical features such as instrument type and ionization mode. These processed attributes are then concatenated into a unified, dense context vector to capture the specific conditions influencing fragmentation patterns..


\textbf{Embedder} Distinct from static featurization, this module serves as the primary neural architecture that learns to map processed chemical inputs into fixed-dimensional latent representations. It acts as the core model backbone rather than merely fetching pre-computed embeddings, it introduces critical inductive biases into the representation learning process. FlexMS implements a diverse suite of embedders—ranging from traditional MLP-based architectures to sequence-based models and graph neural networks (GNNs).

\textbf{Predictor} \quad Integrating the molecular representation and metadata context, this module acts as a decoder to generate the final mass spectrum. It is uniquely challenging because it requires mapping dense latent states into highly sparse $m/z$ peak intensity distributions constrained by complex physical fragmentation rules. FlexMS adapts three distinct predictor paradigms: MassFormer, NEIMS, and MolMS, representing varying levels of model capacity and architectural complexity.

Detailed explanations of the featurization processes, network architecture detail, and the metadata embedding mechanisms are provided in the \textit{Appendix}~\ref{sec:evaluation_metric}.

\subsection{Evaluation Metrics}
To comprehensively evaluate model performance, FlexMS supports a wide array of metrics tailored for different downstream objectives. For generative spectrum prediction, it includes \textbf{Cosine Similarity}, \textbf{Jensen-Shannon Similarity} and \textbf{Spectral Coverage} to assess peak pattern resemblance and probability distribution alignment. For practical compound identification (retrieval tasks), FlexMS implements ranking-based metrics, including \textbf{Absolute / Normalized Rank}, and \textbf{Top-K / Top-K\% Accuracy}. Mathematical formulations for all metrics are detailed in the \textit{Appendix}~\ref{app:methods_details}.

\subsection{Supporting Datasets}
FlexMS standardizes data loading and processing for major public mass spectrometry resources. Supported datasets include \textbf{GNPS} (a large-scale, crowdsourced library)~\cite{wang2016sharing}, \textbf{MassBank} (high-resolution reference spectra)~\cite{horai2010massbank}, \textbf{MassSpecGym} (a standardized ML benchmark)~\cite{bushuiev2024massspecgym} and \textbf{NPLIB1} (a curated natural products collection)~\cite{duhrkop2021systematic}. Furthermore, FlexMS supports challenge-based datasets like \textbf{CASMI} 2016 and 2022~\cite{schymanski2017critical} to evaluate real-world compound retrieval. Detailed dataset statistics and split evaluations are available in the \textit{Appendix}~\ref{app:datasets_details}.

\section{Results}
\label{sec:results}

\subsection{Dataset Characterization in Mass Spectrum Prediction}

\textbf{Structural Diversity quantifies the distributional shifts of splits.} In the rapidly evolving field of mass spectrometry, the choice of datasets plays a pivotal role in ensuring robust generalization. A key aspect of molecular datasets is structural diversity, which directly impacts the difficulty of prediction tasks~\cite{mason1999diversity}. We evaluated structural similarity using Tanimoto coefficients based on Morgan fingerprints~\cite{rogers2010extended}, comparing scaffold splits against random splits (methods are in \textit{Appendix}~\ref{sub:kolmogorov}.

As shown in Table~\ref{tab:stats-metrics}, despite similar mean Tanimoto similarities across splits, two-sample Kolmogorov-Smirnov (KS) tests reveal fundamental differences in data leakage. Scaffold splits consistently produce elevated KS statistics with highly significant $p$-values, confirming substantial distributional shifts that rigorously challenge model generalization~\cite{schuffenhauer2007scaffold} (with $\log(\mathrm{pval}) < -50$). To provide independent validation from a spectral perspective, we also analyzed the distribution of spectral entropy, which corroborates these structural findings (details are in \textit{Appendix}~\ref{app:spectral_entropy}).

\begin{table}[htbp]
\centering
\caption{Comparison of structural similarity metrics across different datasets and splits. MB denotes MassBank, and MSG denotes MassSpecGym. The suffixes "-R" and "-S" 
indicate random and scaffold splits. NPLIB1-0, 1, and 2 correspond to predefined 
splits of the NPLIB1 dataset. Tanimoto is the mean cross-set Tanimoto similarity. KS-stat is the Kolmogorov-Smirnov statistic comparing within-set vs cross-set similarity distributions. Log KS-pval is $\log(\text{p-value})$ from the KS test.}
\label{tab:stats-metrics}
\resizebox{\textwidth}{!}{
\begin{tabular}{lcccccccc}
\toprule
\textbf{Train-Val} & \textbf{GNPS-S} & \textbf{GNPS-R} & \textbf{MB-S} & \textbf{MB-R} & \textbf{MSG} & \textbf{NPLIB1-0} & \textbf{NPLIB1-1} & \textbf{NPLIB1-2} \\
\midrule
Tanimoto & 0.1059  & 0.1056 & 0.0795  & 0.0875 & 0.1002 & 0.1032 & 0.1052 & 0.1055 \\
KS-stat       & 0.0528  & 0.0024 & 0.0831  & 0.0044 & 0.0345 & 0.0188 & 0.0051 & 0.0039 \\
Log KS-pval   & \textbf{-120.64} & -0.03  & \textbf{-299.87} & -0.53  & \textbf{-51.32} & \textbf{-14.97} & -0.84  & -0.36 \\
\midrule
\textbf{Train-Test} & \textbf{GNPS-S} & \textbf{GNPS-R} & \textbf{MB-S} & \textbf{MB-R} & \textbf{MSG} & \textbf{NPLIB1-0} & \textbf{NPLIB1-1} & \textbf{NPLIB1-2} \\
\midrule
Tanimoto & 0.1045  & 0.1051 & 0.0788  & 0.0871 & 0.1005 & 0.1053 & 0.1052 & 0.1040 \\
KS-stat       & 0.0326  & 0.0050 & 0.0794  & 0.0094 & 0.1002 & 0.0033 & 0.0020 & 0.0232 \\
Log KS-pval   & \textbf{-45.74}  & -0.079 & \textbf{-274.18} & -3.51  & \textbf{-61.88} & -0.07  & -0.01  & \textbf{-6.23} \\
\bottomrule
\end{tabular}
}
\end{table}

\subsection{Empirical Study on Embedder and Prediction Models}
\label{sec:empirical_study}

\textbf{GFv2 and Massformer has the highest performance.} We systematically evaluated diverse combinations of embedder and predictor architectures. Table \ref{tab:embedder_predictor_performance_transposed} shows the result. Among the tested embedders, GraphormerV2 (GFv2)~\cite{yang2021graphformers} consistently achieved the highest performance across all predictor methods and resolutions, demonstrating superior capability in molecular representation. Graph-based methods showed competitive performance, while transformer-based and CNN methods exhibited moderate results. Among the predictors, MassFormer~\cite{young2024tandem} achieved the highest overall metrics, followed by NEIMS~\cite{wei2019rapid} and MolMS~\cite{hong20233dmolms}. These consistent performance hierarchies are not limited to MassSpecGym (\textit{Appendix}~\ref{app:additional_results} and Figure~\ref{fig:benchmark}).

Critically, our results highlight strong synergistic relationships between specific molecular representation methods and prediction architectures. For instance, the optimal combination was the GFv2 embedder paired with the MassFormer predictor. MolMS predictors demonstrate superior performance when paired with GNN-based embedders, as both components operate within a graph-theoretic framework that naturally captures spatial relationships~\cite{kipf2016semi}, and NEIMS shows enhanced effectiveness when combined with MLP-based embedders.

\begin{table}[t]
\centering
\caption{Performance of different embedder-predictor combinations on the MassSpecGym dataset. Best, second best, and third best results in each row are denoted by \textbf{bold}, \underline{\underline{doubleline}}, and \underline{underline}, respectively. "Trans", "AttnFP", "DL", "AllDes", "CBTa" denotes Transformers, AttentiveFP, Daylight, All-Descriptors MLP and ChemBERTa. In predictor, "Mass" denotes MassFormers. }
\label{tab:embedder_predictor_performance_transposed}
\resizebox{\textwidth}{!}{%
\begin{tabular}{l*{11}{>{\centering\arraybackslash}p{0.75cm}}}
\toprule
\textbf{Pred} & \textbf{CNN} & \textbf{GCN} & \textbf{GIN} & \textbf{GAT} & \textbf{Trans} & \textbf{ESPF} & \textbf{DL} & \textbf{Attn} & \textbf{AllDes} & \textbf{CBTa} & \textbf{GFv2} \\
\midrule
\multicolumn{12}{l}{Cosine Similarity (Cos Sim)} \\
\midrule
Mass       & 0.316 & 0.306 & \underline{0.329} & 0.317 & 0.310 & 0.310 & 0.324 & 0.323 & \underline{\underline{0.332}} & 0.317 & \textbf{0.357} \\
NEIMS      & 0.303 & 0.302 & \underline{\underline{0.324}} & 0.304 & 0.305 & 0.297 & 0.308 & 0.314 & \underline{0.319} & 0.303 & \textbf{0.354} \\
MolMS      & 0.226 & 0.240 & \underline{0.247} & 0.240 & 0.224 & 0.212 & 0.229 & \underline{\underline{0.248}} & 0.241 & 0.233 & \textbf{0.271} \\
\midrule
\multicolumn{12}{l}{Jensen-Shannon Similarity (JS Sim)} \\
\midrule
Mass       & 0.477 & 0.484 & \underline{\underline{0.494}} & 0.483 & 0.471 & 0.475 & 0.481 & 0.489 & \underline{0.491} & 0.480 & \textbf{0.516} \\
NEIMS      & 0.455 & \underline{0.468} & 0.461 & 0.458 & 0.456 & 0.454 & 0.462 & 0.461 & \underline{\underline{0.470}} & 0.460 & \textbf{0.488} \\
MolMS      & 0.433 & 0.440 & \underline{0.442} & 0.439 & 0.430 & 0.430 & 0.436 & 0.442 & \underline{\underline{0.446}} & 0.436 & \textbf{0.457} \\
\bottomrule
\end{tabular}%
}
\end{table}

\textbf{High cost of GFv2 makes efficient alternatives more practical under constraints.} As detailed in Table~\ref{tab:computational}, GFv2 requires substantially more time and more GPU memory. Training GFv2 on large datasets like GNPS can require several days even on A100 GPU. In resource-constrained environments, models like GIN or simple MLPs provide a much better trade-off. For memory-constrained scenarios or applications requiring high-throughput inference, we highly recommend the GIN or All-descriptors MLP.

\begin{table}[htbp]
\centering
\caption{Computational efficiency comparison across model architectures. Time (seconds) and peak GPU memory (MB) per forward pass (in training) are measured during training with a batch size of 512. Asterisks (*) indicate pretrained models.}
\label{tab:computational}
\resizebox{\textwidth}{!}{
\begin{tabular}{l*{11}{>{\centering\arraybackslash}p{0.75cm}}}
\toprule
\textbf{Time} & \textbf{CNN} & \textbf{GCN} & \textbf{GIN} & \textbf{GAT} & \textbf{Trans} & \textbf{ESPF} & \textbf{DL} & \textbf{Attn} & \textbf{AllDes} & \textbf{CBTa$^*$} & \textbf{GFv2$^*$} \\
\midrule
Mass & 0.532 & 0.656 & 0.658 & 0.823 & 0.270 & 0.395 & 0.394 & 0.690 & 0.351 & 3.493 & 123.5 \\
NEIMS      & 0.471 & 0.581 & 0.564 & 0.594 & 0.212 & 0.324 & 0.355 & 0.527 & 0.346 & 3.432 & 121.9 \\
MolMS      & 0.473 & 0.467 & 0.498 & 0.538 & 0.202 & 0.318 & 0.312 & 0.523 & 0.344 & 3.415 & 122.6 \\
\midrule
\textbf{Mem} & \textbf{CNN} & \textbf{GCN} & \textbf{GIN} & \textbf{GAT} & \textbf{Trans} & \textbf{ESPF} & \textbf{DL} & \textbf{Attn} & \textbf{AllDes} & \textbf{CBTa$^*$} & \textbf{GFv2$^*$} \\
\midrule
Mass & 228.1 & 341.4 & 428.7 & 296.4 & 2882 & 538.1 & 672.9 & 1201  & 781.5 & 982.2 & 24403 \\
NEIMS      & 122.8 & 112.1 & 144.2 & 122.0 & 2554 & 135.2 & 165.2 & 746.7 & 209.7 & 476.7 & 23105 \\
MolMS      & 271.9 & 264.0 & 299.7 & 272.5 & 2717 & 290.7 & 323.7 & 906.7 & 375.8 & 643.9 & 23587 \\
\bottomrule
\end{tabular}
}
\end{table}

\subsection{Impact of Hyperparameters, Data Sparsity and Pretrain}
\label{sec:hyperparameters}

To provide actionable guidance for model deployment in resource-constrained environments, we investigated the impact of training data sparsity and learning rate variations on prediction performance. 

\textbf{Data Sparsity will influence the best model.} We subsampled the MassSpecGym dataset into 25\%, 50\%, and 100\% scaffold fractions. As shown in Table~\ref{tab:hyperparameters_cos} and ~\ref{tab:hyperparameters_js}, models with complex inductive biases experienced severe performance degradation under the 25\% regime, indicating a high propensity for overfitting. In contrast, fingerprint-based MLPs and models leveraging pre-trained backbones maintained robust performance even at low data volumes, though CNNs eventually plateaued as data scaled to 100\% where GNNs could fully leverage their structural inductive biases.

\begin{table*}[t]
\centering
\caption{Impact of data sparsity on embedder performance, evaluated on Cosine Similarity ($\uparrow$).}
\label{tab:hyperparameters_cos}
\resizebox{\textwidth}{!}{%
\begin{tabular}{l*{11}{>{\centering\arraybackslash}p{0.75cm}}}
\toprule
\textbf{Data} & \textbf{CNN} & \textbf{GCN} & \textbf{GIN} & \textbf{GAT} & \textbf{Trans} & \textbf{ESPF} & \textbf{DL} & \textbf{Attn} & \textbf{AllDes} & \textbf{CBTa} & \textbf{GFv2} \\
\midrule
25\% & \textbf{0.254} & 0.185 & 0.229 & 0.236 & 0.218 & \underline{\underline{0.251}} & 0.244 & 0.225 & \underline{0.245} & 0.244 & 0.241 \\
50\% & \underline{\underline{0.293}} & 0.236 & 0.281 & 0.288 & 0.266 & 0.280 & 0.292 & 0.285 & 0.286 & \underline{0.290} & \textbf{0.306} \\
100\% & 0.316 & 0.306 & \underline{0.329} & 0.317 & 0.310 & 0.310 & 0.324 & 0.323 & \underline{\underline{0.332}} & 0.317 & \textbf{0.357} \\
\bottomrule
\end{tabular}%
}
\end{table*}

\textbf{Optimal learning rates depend strongly on dataset noise and curation quality.} Table~\ref{tab:lr_all_datasets} shows the result. For heterogeneous datasets such as MassBank, lower learning rates ($10^{-4}$) are preferable for navigating complex optimization landscapes. In contrast, highly curated datasets such as NPLIB1 offer smoother loss surfaces, enabling faster convergence with higher learning rates ($10^{-3}$).
\textit{Appendix} Figure~\ref{fig:FlexMS-LR-supp} provides the broader multi-metric comparison across datasets, predictors, and resolutions, while \textit{Appendix} Figure~\ref{fig:FlexMS-LR-retrieval} shows the corresponding retrieval-side sensitivity.

\begin{table}[htbp]
\centering
\caption{Impact of learning rate on embedder performance across three datasets (1 Da resolution). Performance is evaluated via Cosine Similarity.}
\label{tab:lr_all_datasets}
\resizebox{\textwidth}{!}{
\begin{tabular}{lc *{10}{>{\centering\arraybackslash}p{0.7cm}}}
\toprule
Method & LR & \multicolumn{10}{c}{Embedder} \\
\cmidrule{3-12}
& & CNN & GCN & GIN & GAT & Trans & ESPF & DL & Attn & AllDes & CBTa \\
\midrule
\multicolumn{12}{l}{\textbf{MB}} \\
\midrule
\multirow{2}{*}{MF} & $10^{-3}$ & 0.534 & 0.504 & 0.516 & 0.536 & 0.478 & \textbf{0.460} & 0.445 & \textbf{0.522} & \textbf{0.468} & 0.521 \\
 & $10^{-4}$ & \textbf{0.546} & \textbf{0.531} & \textbf{0.538} & \textbf{0.544} & \textbf{0.497} & 0.455 & \textbf{0.461} & 0.521 & 0.467 & \textbf{0.537} \\
\cmidrule{1-12}
\multirow{2}{*}{NEIMS} & $10^{-3}$ & 0.454 & \textbf{0.506} & 0.498 & \textbf{0.511} & 0.445 & 0.408 & 0.442 & 0.398 & 0.405 & 0.461 \\
 & $10^{-4}$ & \textbf{0.488} & 0.503 & \textbf{0.519} & 0.508 & \textbf{0.477} & \textbf{0.424} & \textbf{0.456} & \textbf{0.469} & \textbf{0.451} & \textbf{0.492} \\
\cmidrule{1-12}
\multirow{2}{*}{MolMS} & $10^{-3}$ & 0.433 & 0.445 & 0.451 & \textbf{0.491} & 0.385 & 0.348 & 0.362 & 0.400 & 0.371 & 0.420 \\
 & $10^{-4}$ & \textbf{0.449} & \textbf{0.451} & \textbf{0.474} & 0.465 & \textbf{0.410} & \textbf{0.350} & \textbf{0.371} & \textbf{0.434} & \textbf{0.371} & \textbf{0.451} \\
\midrule
\multicolumn{12}{l}{\textbf{MSG}} \\
\midrule
\multirow{2}{*}{MF} & $10^{-3}$ & 0.301 & 0.312 & 0.328 & 0.316 & \textbf{0.319} & \textbf{0.311} & 0.308 & \textbf{0.324} & \textbf{0.316} & 0.326 \\
 & $10^{-4}$ & \textbf{0.306} & \textbf{0.324} & \textbf{0.332} & \textbf{0.323} & 0.317 & 0.310 & \textbf{0.310} & 0.317 & 0.316 & \textbf{0.329} \\
\cmidrule{1-12}
\multirow{2}{*}{NEIMS} & $10^{-3}$ & 0.292 & 0.298 & 0.316 & 0.311 & 0.296 & \textbf{0.297} & \textbf{0.306} & 0.296 & 0.300 & 0.320 \\
 & $10^{-4}$ & \textbf{0.302} & \textbf{0.308} & \textbf{0.319} & \textbf{0.314} & \textbf{0.303} & 0.297 & 0.305 & \textbf{0.304} & \textbf{0.303} & \textbf{0.324} \\
\cmidrule{1-12}
\multirow{2}{*}{MolMS} & $10^{-3}$ & 0.233 & 0.221 & 0.236 & 0.239 & 0.232 & \textbf{0.214} & 0.222 & 0.239 & \textbf{0.227} & 0.233 \\
 & $10^{-4}$ & \textbf{0.240} & \textbf{0.229} & \textbf{0.241} & \textbf{0.248} & \textbf{0.233} & 0.212 & \textbf{0.224} & \textbf{0.240} & 0.226 & \textbf{0.247} \\
\midrule
\multicolumn{12}{l}{\textbf{NPLIB1}} \\
\midrule
\multirow{2}{*}{MF} & $10^{-3}$ & 0.557 & 0.529 & 0.561 & 0.549 & \textbf{0.539} & \textbf{0.526} & 0.530 & \textbf{0.561} & 0.527 & \textbf{0.560} \\
 & $10^{-4}$ & \textbf{0.560} & \textbf{0.545} & \textbf{0.570} & \textbf{0.557} & 0.534 & 0.524 & \textbf{0.538} & 0.558 & \textbf{0.533} & 0.550 \\
\cmidrule{1-12}
\multirow{2}{*}{NEIMS} & $10^{-3}$ & \textbf{0.552} & \textbf{0.543} & \textbf{0.559} & \textbf{0.552} & \textbf{0.545} & \textbf{0.517} & \textbf{0.538} & \textbf{0.551} & \textbf{0.510} & \textbf{0.555} \\
 & $10^{-4}$ & 0.512 & 0.526 & 0.526 & 0.525 & 0.508 & 0.485 & 0.513 & 0.491 & 0.486 & 0.503 \\
\cmidrule{1-12}
\multirow{2}{*}{MolMS} & $10^{-3}$ & \textbf{0.512} & \textbf{0.487} & 0.500 & \textbf{0.517} & \textbf{0.501} & \textbf{0.481} & 0.452 & \textbf{0.510} & \textbf{0.475} & 0.489 \\
 & $10^{-4}$ & 0.498 & 0.485 & \textbf{0.526} & 0.513 & 0.497 & 0.474 & \textbf{0.503} & 0.476 & 0.473 & \textbf{0.490} \\
\bottomrule
\end{tabular}
}
\end{table}

\textbf{Molecular Pretraining matters.} The lack of experimentally annotated MS/MS spectra makes self-supervised pretraining a highly attractive strategy. We assessed the performance of randomly initialized embedders against their pretrained counterparts across different datasets and split methods, and we choose MoleBERT because of its tiny but calibratedly designed technique \citep{xia2023mole}.
\label{sec:pretraining}

Our results in Table~\ref{tab:res1_summary} indicate that pretraining provides only marginal improvements on random splits, but more consistent and often larger gains on scaffold splits. \textbf{This suggests that self-supervised pretraining on large, unlabeled molecular corpora helps the embedder capture general chemical regularities, thereby improving generalization to unseen molecular scaffolds.} Furthermore, pretrained models converged faster during fine-tuning on the target mass spectrum datasets. A broader cross-dataset and multi-metric comparison is provided in \textit{Appendix} Figure~\ref{fig:FlexMS-pretrained}.

\subsection{Cross-Domain Transfer Learning Analysis}
\label{sec:transfer_learning}

Real-world applications often require applying spectrum-prediction models to spectra acquired under different experimental conditions from those seen during training. To evaluate robustness under such domain shifts, cross-domain transfer experiments were conducted on MassSpecGym across two axes: ionization mode ([M+H]$^+$ vs. [M+Na]$^+$) and instrument type (Orbitrap vs. QTOF).

The results in Table~\ref{tab:transfer_combined} reveal asymmetric behavior that cannot be explained by sample size alone. MassSpecGym is strongly imbalanced across these domains: approximately 85\% of spectra are [M+H]$^+$ and 15\% are [M+Na]$^+$. \textbf{Despite much smaller source domain, models trained on [M+Na]$^+$ spectra often transfer better to [M+H]$^+$ spectra than the reverse direction.} \textbf{Previous papers demonstrated that sodium-adducted ions exhibit markedly different fragmentation behavior~\cite{liu2025adduct}}, with a low similarity to [M+H]+ due to charge-remote fragmentation and coordination-based stabilization mechanisms. We hypothesize that training on these highly complex, remote-cleavage patterns forces the model to learn deeper and more robust topological representations of the molecular graph, which acts as a strong regularization mechanism,

A similar asymmetry appears across instrument, where roughly 75\% are on Orbitrap instruments and 25\% on QTOF. \textbf{Models trained on high-resolution Orbitrap spectra generally transfer better to QTOF spectra than QTOF-trained models transfer to Orbitrap spectra, which is consistent with the higher mass accuracy and more spectral detail of Orbitrap measurements~\cite{deschamps2023advances}.}

These results indicate that domain coverage should be evaluated both on amount of spectra, the information content and transferability of the source domain. Extended transfer diagnostics, including direction-wise visual comparisons and critical-difference analysis across embedders, are provided in the \textit{Appendix}~\ref{app:cross_domain}. Same-domain simulations are also detailed in \textit{Appendix}~\ref{app:same_domain}.

\begin{table}[htbp]
\centering
\caption{Performance comparison at 1 Bin Resolution with or without pretrained MoleBERT model.}
\label{tab:res1_summary}
\begin{tabular}{l cc cc cc cc}
\toprule
\multirow{2}{*}{\textbf{Method}} & \multicolumn{2}{c}{\textbf{MB (Random)}} & \multicolumn{2}{c}{\textbf{MB (Scaffold)}} & \multicolumn{2}{c}{\textbf{MassSpecGym}} & \multicolumn{2}{c}{\textbf{NPLIB1}} \\
\cmidrule(lr){2-3} \cmidrule(lr){4-5} \cmidrule(lr){6-7} \cmidrule(lr){8-9}
& Pretrain & w/o Pre & Pretrain & w/o Pre & Pretrain & w/o Pre & Pretrain & w/o Pre \\
\midrule
Mass & \textbf{0.5363} & 0.5202 & \textbf{0.4124} & 0.3844 & \textbf{0.3406} & 0.3138 & \textbf{0.5939} & 0.5337 \\
NEIMS      & \textbf{0.5201} & 0.5164 & \textbf{0.3950} & 0.3792 & \textbf{0.3469} & 0.3127 & \textbf{0.5603} & 0.5374 \\
MolMS      & \textbf{0.4731} & 0.4508 & \textbf{0.3389} & 0.3162 & \textbf{0.2467} & 0.2215 & \textbf{0.5265} & 0.4693 \\
\bottomrule
\end{tabular}
\end{table}
\begin{table}[htbp]
\centering
\caption{Cross-domain transfer learning performance. Transfer directions are denoted using: Q (QTOF) and O (Orbitrap) for instrument types; H ([M+H]$^+$) and Na ([M+Na]$^+$) for ionization modes.}
\label{tab:transfer_combined}
\resizebox{\textwidth}{!}{%
\begin{tabular}{lc*{11}{>{\centering\arraybackslash}p{0.74cm}}}
\toprule
\textbf{Pred} & \textbf{Dir.} & \textbf{CNN} & \textbf{GCN} & \textbf{GIN} & \textbf{GAT} & \textbf{Trans} & \textbf{ESPF} & \textbf{DL} & \textbf{Attn} & \textbf{AllDes} & \textbf{CBTa} & \textbf{GFv2} \\
\midrule
\multicolumn{13}{l}{Instrument Types} \\
\midrule
\multirow{2}{*}{Mass}  
  & Q$\to$O & 0.255 & 0.228 & 0.268 & 0.274 & 0.262 & 0.200 & 0.244 & 0.267 & 0.273 & 0.216 & \textbf{0.285} \\
 & O$\to$Q & 0.303 & 0.294 & 0.323 & 0.332 & 0.314 & 0.184 & 0.324 & 0.315 & 0.311 & 0.303 & \textbf{0.345} \\
\cmidrule{1-13}
\multirow{2}{*}{NEIMS}  
  & Q$\to$O & 0.260 & 0.242 & \textbf{0.276} & 0.243 & 0.259 & 0.252 & 0.245 & 0.263 & 0.252 & 0.251 & 0.269 \\
 & O$\to$Q & 0.318 & 0.289 & 0.326 & 0.316 & 0.316 & 0.312 & 0.317 & 0.324 & 0.328 & 0.320 & \textbf{0.351} \\
\cmidrule{1-13}
\multirow{2}{*}{MolMS}  
  & Q$\to$O & 0.175 & 0.190 & 0.179 & 0.192 & 0.179 & 0.164 & 0.156 & 0.198 & 0.180 & 0.152 & \textbf{0.242} \\
 & O$\to$Q & 0.254 & 0.266 & 0.240 & 0.251 & 0.263 & 0.194 & 0.131 & 0.267 & 0.211 & 0.196 & \textbf{0.302} \\
\midrule
\multicolumn{13}{l}{Ionization Modes} \\
\midrule
\multirow{2}{*}{Mass}  
  & H$\to$Na & 0.131 & 0.176 & 0.203 & 0.209 & \textbf{0.212} & 0.180 & 0.183 & 0.204 & 0.166 & 0.179 & 0.171 \\
 & Na$\to$H & 0.250 & 0.196 & \textbf{0.273} & 0.258 & 0.210 & 0.252 & 0.245 & 0.269 & 0.260 & 0.250 & 0.226 \\
\cmidrule{1-13}
\multirow{2}{*}{NEIMS}  
  & H$\to$Na & 0.219 & 0.176 & 0.227 & 0.212 & 0.228 & 0.221 & 0.222 & \textbf{0.234} & 0.211 & 0.216 & 0.227 \\
 & Na$\to$H & 0.241 & 0.209 & 0.239 & 0.235 & \textbf{0.242} & 0.224 & 0.233 & 0.231 & 0.226 & 0.236 & 0.233 \\
\cmidrule{1-13}
\multirow{2}{*}{MolMS}  
  & H$\to$Na & 0.094 & 0.103 & 0.106 & 0.106 & \textbf{0.117} & 0.090 & 0.081 & 0.110 & 0.085 & 0.093 & 0.107 \\
 & Na$\to$H & 0.206 & 0.205 & 0.205 & 0.211 & 0.211 & 0.204 & 0.212 & \textbf{0.213} & 0.208 & 0.199 & 0.212 \\
\bottomrule
\end{tabular}%
}
\end{table}

\subsection{Retrieval Benchmarks for Practical Identification}
\label{sec:retrieval_benchmarks}

While generative metrics such as cosine similarity and Jensen--Shannon similarity assess the fidelity of predicted spectra, they do not directly measure utility in compound identification. FlexMS therefore includes a retrieval benchmark designed to evaluate whether predicted spectra are sufficiently discriminative to rank the correct molecular structure near the top of a realistic candidate list.

The retrieval benchmark follows CASMI-style identification settings on both CASMI 2016 and 2022. Candidate sets are generated from mass-based searches and then filtered for structural validity, after which each candidate molecule is scored by similarity between its predicted spectrum under the query metadata and the observed query spectrum. Full protocol details and candidate-pool statistics and metric definitions are provided in the \textit{Appendix}~\ref{app:retrieval_setting}.

Models with strong reconstruction metrics generally remain competitive in retrieval, but the relationship is not perfectly linear. Still, GNN-based embedders such as GFv2 achieve the strongest normalized-rank performance across predictors in CASMI 2016 (Table~\ref{tab:retrieval_metrics_combined}), indicating better discrimination ability. Specifically, we discuss GCN in \textit{Appendix}~\ref{app:extended_conclusions}. By contrast, simpler descriptor-based baselines can remain competitive on average reconstruction scores while producing less selective rankings. Extended evaluations on the CASMI 2022 dataset and respective analysis further corroborate these trends (see \textit{Appendix}~\ref{app:retrieval_setting} and Table~\ref{tab:retrieval_metrics_casmi22}).

\begin{table}[ht]
\centering
\caption{Retrieval benchmark performance evaluated on the CASMI 2016 dataset. Metrics include Recall@1 (R@1), Recall@10 (R@10), and Normalized Rank (NormR $\downarrow$) across candidate pools. Best, second best, and third best results are denoted by \textbf{bold}, \underline{\underline{doubleline}}, \underline{underline}, respectively.}
\label{tab:retrieval_metrics_combined}
\resizebox{\textwidth}{!}{
\begin{tabular}{lccccccccc}
\toprule
 & \multicolumn{3}{c}{\textbf{MassFormer}} & \multicolumn{3}{c}{\textbf{NEIMS}} & \multicolumn{3}{c}{\textbf{MolMS}} \\
\cmidrule(lr){2-4} \cmidrule(lr){5-7} \cmidrule(lr){8-10}
& \textbf{R@1} & \textbf{R@10} & \textbf{NormR$\downarrow$} & \textbf{R@1} & \textbf{R@10} & \textbf{NormR$\downarrow$} & \textbf{R@1} & \textbf{R@10} & \textbf{NormR$\downarrow$} \\
\midrule
CNN          & 0.0354 & 0.1869 & 0.3283 & 0.0202 & 0.1313 & 0.3385 & 0.0152 & 0.0758 & 0.3957 \\
GCN          & \underline{\underline{0.1667}} & 0.5455 & 0.1470 & \underline{0.0808} & \underline{0.5354} & \underline{0.1312} & \underline{0.1465} & \underline{\underline{0.5303}} & 0.1969 \\
GIN          & 0.1061 & 0.4949 & \underline{0.1381} & \underline{0.0808} & 0.4798 & 0.1457 & 0.0960 & 0.4394 & 0.1838 \\
GAT          & 0.0859 & 0.4848 & 0.1522 & 0.0707 & 0.3889 & 0.1756 & 0.0354 & 0.3030 & 0.2077 \\
Trans        & 0.0354 & 0.1212 & 0.3699 & 0.0253 & 0.1364 & 0.3523 & 0.0051 & 0.0707 & 0.4284 \\
ESPF         & 0.0354 & 0.1212 & 0.3809 & 0.0101 & 0.1010 & 0.3636 & 0.0051 & 0.0909 & 0.3860 \\
DL           & \underline{0.1515} & 0.4495 & 0.1988 & \underline{\underline{0.1818}} & \underline{\underline{0.5505}} & 0.1487 & \underline{0.1465} & 0.5051 & \underline{0.1611} \\
Attn       & 0.1313 & \underline{\underline{0.6111}} & \underline{\underline{0.1286}} & \underline{0.0808} & 0.5253 & \underline{\underline{0.1267}} & 0.1263 & \underline{0.5909} & \textbf{0.1566} \\
AllDes       & \textbf{0.1970} & \underline{0.5808} & 0.1696 & \textbf{0.2071} & \underline{\underline{0.5505}} & 0.1324 & \underline{0.1364} & 0.4394 & 0.1852 \\
CBTa          & 0.0354 & 0.1162 & 0.3394 & 0.0303 & 0.1465 & 0.3135 & 0.0101 & 0.1010 & 0.4141 \\
GFv2         & \underline{0.1515} & \textbf{0.6667} & \textbf{0.1022} & \textbf{0.2071} & \textbf{0.6414} & \textbf{0.1133} & \textbf{0.1717} & \textbf{0.6061} & \underline{\underline{0.1574}} \\
\bottomrule
\end{tabular}
}
\end{table}

\section{Conclusion and Limitations}
\label{sec:conclusion}

In this work, we introduced FlexMS, a modular and comprehensive benchmarking framework designed to standardize the evaluation of deep learning-based mass spectrum prediction tools. By decoupling molecular prediction pipeline and systematically evaluating them across diverse public metabolomics resources, we demonstrated the model accuracy is highly contingent on the underlying evaluation protocol. Our empirical results establish that while GraphormerV2 paired with MassFormer consistently achieves the strongest performance under our unified protocol, this performance requires substantial computational overhead. By rigorously evaluating cross-domain robustness and validating distribution shifts through structural and spectral metrics, we demonstrate that realistic problem settings, such as scaffold splits and instrument transfer, are vital for guiding generalizable and practical model selection.

While our framework systematically evaluates a wide range of representative components, it does not currently cover all the latest state-of-the-art methods or advanced training tricks in this field. However, the contribution of FlexMS is not merely identifying a single winning architecture but rather providing a reproducible ecosystem. This framework empowers researchers to move beyond entangled comparisons, enabling them to rigorously benchmark future innovations and identify the most practical operating points.




\clearpage
\bibliographystyle{unsrt}
\bibliography{sn-bibliography}

\clearpage
\appendix

\newpage

\renewcommand{\thefigure}{S\arabic{figure}}
\renewcommand{\thetable}{S\arabic{table}}
\renewcommand{\theHfigure}{supp.\arabic{figure}}
\renewcommand{\theHtable}{supp.\arabic{table}}
\setcounter{figure}{0}
\setcounter{table}{0}

\section{Supplementary Material}
\label{sec:material}

\subsection{Complementary Validation via Spectral Entropy Analysis}
\label{app:spectral_entropy}

To provide independent validation from the spectral perspective, we analyzed the distribution of spectral entropy across dataset splits. Spectral entropy~\cite{li2021spectral}, defined as $H = -\sum_{i} p_i \ln(p_i)$ where $p_i$ represents the normalized peak intensity, quantifies the complexity of fragmentation patterns in mass spectra. While the Tanimoto-based analysis examines structural similarity at the molecular input level, spectral entropy captures differences in fragmentation behavior at the output level.

As shown in Table~\ref{tab:entropy-metrics}, the mean spectral entropy varies across datasets. GNPS, MassBank, and MassSpecGym exhibit similar entropy levels ($H \approx 1.6$--$1.7$), whereas NPLIB1 shows notably higher entropy ($H \approx 2.8$), reflecting the complex fragmentation patterns characteristic of natural products~\cite{li2021spectral}. More importantly, scaffold splits exhibit substantially larger distributional shifts in spectral entropy compared to random splits. For GNPS, the KS statistic under scaffold split is 4.3 times higher than random split (0.011) with significant p-values, and MassBank shows a 7.1-fold increase in KS statistics for scaffold versus random splits similarly. 

\begin{table}[ht]
\centering
\caption{Spectral entropy distribution analysis across dataset splits. Mean Entropy denotes the average Shannon entropy (in nats) of training set spectra. KS-stat and Log KS-pval report the Kolmogorov-Smirnov test statistics and $\log_{10}$-transformed p-values comparing entropy distributions between splits. Bold values indicate highly significant distributional shifts.}
\label{tab:entropy-metrics}
\resizebox{\textwidth}{!}{
\begin{tabular}{lcccccccc}
\toprule
\textbf{Train-Val} & \textbf{GNPS-S} & \textbf{GNPS-R} & \textbf{MB-S} & \textbf{MB-R} & \textbf{MSG} & \textbf{NPLIB1-0} & \textbf{NPLIB1-1} & \textbf{NPLIB1-2} \\
\midrule
Mean Entropy & 1.7275 & 1.7169 & 1.6374 & 1.7027 & 1.7396 & 2.8494 & 2.8486 & 2.8326 \\
KS-stat      & 0.0595 & 0.0054 & 0.1245 & 0.0163 & 0.0247 & 0.0340 & 0.0716 & 0.0586 \\
Log KS-pval  & \textbf{-262.11} & -1.66 & \textbf{-171.78} & -2.13 & -9.02 & -0.44 & -2.67 & -1.76 \\
\midrule
\textbf{Train-Test} & \textbf{GNPS-S} & \textbf{GNPS-R} & \textbf{MB-S} & \textbf{MB-R} & \textbf{MSG} & \textbf{NPLIB1-0} & \textbf{NPLIB1-1} & \textbf{NPLIB1-2} \\
\midrule
Mean Entropy & 1.7275 & 1.7169 & 1.6374 & 1.7027 & 1.7396 & 2.8494 & 2.8486 & 2.8326 \\
KS-stat      & \textbf{0.0492} & 0.0115 & \textbf{0.0995} & 0.0140 & 0.0137 & 0.0386 & 0.0332 & 0.0507 \\
Log KS-pval  & \textbf{-172.84} & -7.94 & \textbf{-88.19} & -1.48 & -2.33 & -1.44 & -1.01 & -2.91 \\
\bottomrule
\end{tabular}
}
\end{table}

\section{Extended Conclusions and Discussion}
\label{app:extended_conclusions}

Beyond the primary findings presented in the main text, our extensive ablation studies within the FlexMS framework provide several nuanced insights into model behaviors under specific operational constraints. 

\paragraph{Data Sparsity and Architecture Choice} 

Our data ablation experiments highlight a critical trade-off between model capacity and data availability. We observed that high-capacity architectures with complex inductive biases, such as deep Graph Neural Networks (GNNs) and standard Transformers, are highly susceptible to overfitting in data-scarce cases. In contrast, fingerprint-based Multi-Layer Perceptrons (MLPs) leverage strong, predefined chemical priors to maintain robust performance, making them the preferred choice when annotated MS/MS data is severely limited. \textbf{Interestingly, 1D-CNNs emerged as highly efficient feature extractors in this extreme low-data regime, even outperforming several fingerprint methods.} By treating SMILES strings as sequences, CNNs efficiently capture local substructure patterns like functional groups, without requiring massive datasets to learn global topological rules.

\begin{table*}[htbp]
\centering
\caption{Impact of data sparsity on embedder performance, evaluated via Jensen-Shannon Similarity.}
\label{tab:hyperparameters_js}
\resizebox{\textwidth}{!}{%
\begin{tabular}{l*{11}{>{\centering\arraybackslash}p{0.75cm}}}
\toprule
\textbf{Data} & \textbf{CNN} & \textbf{GCN} & \textbf{GIN} & \textbf{GAT} & \textbf{Trans} & \textbf{ESPF} & \textbf{DL} & \textbf{Attn} & \textbf{AllDes} & \textbf{CBTa} & \textbf{GFv2} \\
\midrule
25\% & \underline{\underline{0.436}} & 0.412 & 0.430 & 0.431 & 0.425 & \underline{0.435} & 0.435 & 0.427 & \textbf{0.438} & 0.435 & 0.431 \\
50\% & 0.458 & 0.436 & 0.459 & 0.457 & 0.445 & 0.452 & 0.457 & 0.455 & \underline{\underline{0.461}} & \underline{0.460} & \textbf{0.475} \\
100\% & 0.477 & 0.484 & \underline{\underline{0.494}} & 0.483 & 0.471 & 0.475 & 0.481 & 0.489 & \underline{0.491} & 0.480 & \textbf{0.516} \\
\bottomrule
\end{tabular}%
}
\end{table*}

\paragraph{Dataset Noise and the Optimization Landscape} 

\textbf{Our empirical study highlights that hyperparameter sensitivity in binned spectrum prediction is deeply intertwined with the specific characteristics of the training corpus. }The robust preference for a lower learning rate ($10^{-4}$) on MassSpecGym can be attributed to the high variability in spectral quality, unpredictable noise, and diverse fragmentation patterns inherent to crowdsourced repositories or synthetic augmentations. In a multi-label regression context with thousands of $m/z$ bins, this heterogeneity creates a rugged, highly non-convex optimization landscape where aggressive gradient steps easily lead to overshooting. In stark contrast, the NPLIB1 dataset consists of curated, high-quality spectra from natural products with predictable and consistent fragmentation rules. This distinct curation process yields a comparatively smoother optimization landscape, permitting models to utilize a higher learning rate ($10^{-3}$) to achieve faster convergence and leverage dataset regularities more aggressively without destabilizing the training process. \textit{Appendix} ~\ref{learning_rate_pred} shows the experiment.

\paragraph{The GCN Bottleneck}

\textbf{Our multi-task evaluation identifies GCN as an outlier because it remains competitive in retrieval despite poor performance in absolute simulation.} This gap is theoretically rooted in GCN's aggregation operator ($D^{-1/2}AD^{-1/2}$) performing signal smoothing. While this smoothing builds robust latent representations for discriminative retrieval by capturing global structural similarities, it acts as a low-pass filter that erases the local heterogeneity required for generative simulation. However, because retrieval evaluates relative ranking rather than absolute peak matching, this oversmoothing paradoxically provides a stable envelope for predictions.

Even though the generated spectra lack fine-grained fidelity, these structurally distinctive envelopes are sufficient to reliably discriminate the true compound from decoys in the candidate pool. In contrast, other backbones like GIN (using sum aggregation) and GAT (via adaptive attention) preserve the atom-specific variance necessary to map chemical graphs to sparse, high-dimensional spectral outputs.

\section{Detailed Methodology}
\label{app:methods_details}

\subsection{Datasets Details}
\label{app:datasets_details}

To comprehensively evaluate the robustness and generalizability of our framework, we have curated a diverse suite of datasets representing various scenarios encountered in metabolic research . An overview of their scale and benchmark role is provided in Table according to NIPS recommendation~\ref{tab:dataset_summary}~\cite{gebru2021datasheets}.

\paragraph{MassSpecGym Dataset} 
MassSpecGym \cite{bushuiev2024massspecgym} serves as a foundational platform for evaluating model performance in a controlled environment. It comprises 231,000 spectra from 29,000 unique molecules. The dataset was specifically designed for machine learning applications, featuring highly standardized metadata. A key advantage is its rigorous data-splitting procedure based on molecular edit distance, designed to prevent data leakage and robustly assess a model's generalization to novel chemical structures.

\paragraph{GNPS Dataset} 
To test model performance under realistic, large-scale conditions, we incorporate the Global Natural Products Social Molecular Networking (GNPS) library \cite{wang2016sharing}. Containing over 322,000 spectra from more than 16,000 molecules, it is characterized by significant heterogeneity from diverse instruments, laboratories, and experimental protocols.

\paragraph{MassBank Dataset} 
MassBank \cite{horai2010massbank} provides a rich, open-source library of over 62,000 spectra from more than 4,000 small chemical compounds, many of which are high-resolution. Its strict record validation ensures high reliability for metabolomics identification.

\paragraph{NPLIB1 Dataset} 
For more targeted analyses where data quality and balance are critical, we employ the NPLIB1 benchmark dataset \cite{duhrkop2021systematic}. It consists of approximately 8,000 high-quality spectra from 7,000 unique natural products, with a careful curation process that ensures an even distribution across chemical classes.

\paragraph{CASMI Contest} 
Challenges from the Critical Assessment of Small Molecule Identification (CASMI) contest \cite{schymanski2017critical} are used to simulate real-world compound identification. The core task involves correctly identifying the true molecular structure for a given query spectrum from a provided list of candidates.

\begin{table*}[h]
\centering
\small
\begin{tabular}{@{}p{2cm}p{2.5cm}p{8cm}@{}}
\toprule
\textbf{Dataset} & \textbf{Data Size(Approx.)} & \textbf{Key Characteristic} \\
\midrule
\textbf{MassSpecGym} & 231k spectra \newline 29k molecules & Standardized ML benchmark with clean metadata and generalization-aware data splits. It serves as a modern baseline for evaluating deep learning models across diverse chemical spaces. \\
\midrule
\textbf{GNPS} & $>$322k spectra \newline $>$16k molecules & Large-scale, crowdsourced library with significant real-world noise and heterogeneity. \\
\midrule
\textbf{MassBank} & $>$62k spectra \newline $>$4k molecules & The first public repository, offering a rich source of high-resolution spectra. It enforces strict record validation and open data standards to ensure high reliability for metabolomics identification. \\
\midrule
\textbf{NPLIB1 \newline (MIST/Canopus)} & $\sim$8k spectra \newline $\sim$7k molecules & Curated natural products collection with balanced chemical classes and high-quality spectra. It serves as a critical testbed for evaluating model performance on biologically relevant chemical spaces.\\
\midrule
\textbf{CASMI16+22} & 312/208 cases (16) \newline 500 cases (22) & Real-world compound identification task, requiring ranking of candidate structures. \\
\bottomrule
\addlinespace
\end{tabular}
\caption{Overview of Datasets Used for Benchmarking}
\label{tab:dataset_summary}
\end{table*}

\subsection{Retrieval Benchmark Setting}
\label{app:retrieval_setting}

\begin{figure}[htbp]
\centering
\includegraphics[width=\textwidth]{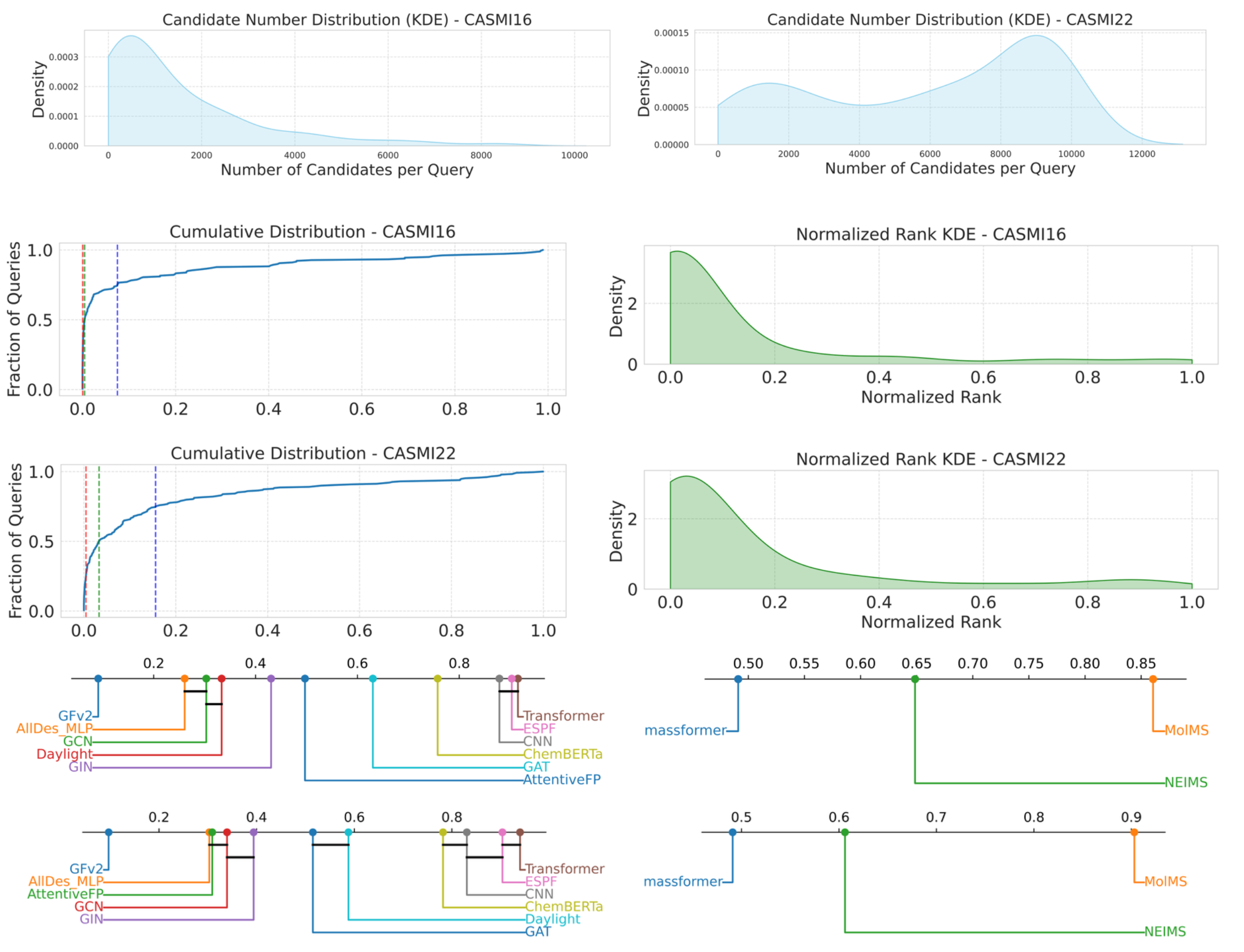}
\caption{\textbf{Upper:} The KDE plot of CASMI compound's candidate number distribution on both CASMI16 and CASMI22 contest. \textbf{Mid:} The cumulative distribution and KDE plot of the normalized-rank performance on GFv2-MassFormerMLP combinations. \textbf{Bottom:} The Critical Difference Diagram of different embedders and predictors on CASMI16 and CASMI22 contest. We compared the performance on normalized rank, so lower value means better rank. }
\label{fig:retrieval}
\end{figure}

\paragraph{Task Background}
Direct spectrum-reconstruction metrics quantify fidelity at the bin level, but practical compound identification is fundamentally a ranking problem. The retrieval benchmark therefore evaluates whether a model can place the correct molecular structure near the top of a candidate list for a given experimental spectrum. This setting follows the identification-oriented formulation used in the Critical Assessment of Small Molecule Identification (CASMI) contests \cite{schymanski2017critical}, where each query is paired with a finite candidate pool and success depends on the rank of the true compound rather than only on reconstruction quality.  Figure ~\ref{fig:retrieval} shows some details.

\paragraph{CASMI 2016 Protocol}
The CASMI 2016 retrieval benchmark uses Orbitrap spectra for 124 compounds after restricting the evaluation to [M+H]+ adducts and excluding charged species. For each query compound, spectra acquired at normalized collision energies (NCEs) of 20, 35, and 50 are combined into a single query representation. Candidate lists are derived from ChemSpider searches using precursor-mass similarity, yielding an average of 1,251 candidates per query after preprocessing. The candidate-count distribution is long-tailed: most compounds have candidate pools on the order of $10^3$, while only a small subset exceeds 4,000 candidates.

\paragraph{CASMI 2022 Protocol}
The CASMI 2022 retrieval benchmark uses Orbitrap spectra for 228 compounds after restricting the evaluation to [M+H]+ and [M+Na]+ adducts and removing compounds with unsupported elements. This restriction keeps the retrieval task aligned with the ionization conditions most consistently represented across the public training corpora. Candidate molecules are generated from PubChem within a 10 ppm precursor-mass tolerance, following the MassFormer retrieval protocol, with at most 10,000 candidates retained per query. After filtering unsupported elements, removing multimolecular compounds, and deduplicating stereoisomers, the resulting candidate pools contain 4,849 molecules per spectrum on average. In contrast to CASMI 2016, the candidate-count distribution is strongly skewed toward the upper cap, indicating a substantially harder low-prior retrieval regime.

\paragraph{Candidate Preparation and Inference}
For each query spectrum $q$, candidate molecules in $C_q$ are matched by precursor $m/z$ and annotated with the query's experimental metadata (e.g., collision energy, instrument type, adduct). After filtering invalid structures, FlexMS predicts a theoretical spectrum for each candidate under these query-specific conditions. The candidates are then ranked by the cosine similarity between their predicted spectra and the observed query spectrum.

\subsection{Data Split Evaluation and Statistical Analysis}

\subsubsection{Molecule Split Setting}
Benchmark conclusions depend strongly on the molecular partition protocol because split design controls the degree of chemical overlap between training and evaluation sets. Two split regimes are considered in this benchmark. A \textit{random split} assigns molecules to training, validation, and test partitions stochastically, which may place structurally related compounds across different subsets. A \textit{scaffold split} instead groups molecules by their Bemis--Murcko scaffolds and assigns all molecules sharing the same scaffold to the same partition, thereby enforcing a stricter structural separation \cite{wu2018moleculenet}.

For GNPS and MassBank, both random and scaffold-based partitions are constructed using an 80\%/10\%/10\% train/validation/test ratio. This paired design enables a direct comparison between easier in-distribution evaluation and harder scaffold-aware evaluation under otherwise matched preprocessing and model settings. For MassSpecGym and NPLIB1, the official benchmark splits are retained in order to preserve the intended evaluation semantics of those resources and to maintain comparability with prior work.

This benchmark design intentionally combines official splits and newly constructed scaffold-aware splits across datasets. The objective is not to force every resource into a single partition scheme, but to expose how model rankings and practical conclusions change when the degree of structural separation varies.
Retrieval benchmarks on CASMI 2016 and CASMI 2022 are treated separately from direct prediction splits, because they are organized around query-specific candidate pools and are evaluated as test sets.

\subsubsection{Kolmogorov-Smirnov Test and Murcko Metric on Morgan fingerprint}
\label{sub:kolmogorov}
Let each molecule $x$ be represented by a binary Morgan fingerprint $\mathbf{f}(x)\in\{0,1\}^d$ on ECFP radius $r$ and bit-length $d$. For two molecules $x, y$, with $a=\|\mathbf{f}(x)\|_1$, $b=\|\mathbf{f}(y)\|_1$, and $c=\mathbf{f}(x)^\top \mathbf{f}(y)$, the Tanimoto similarity is $T(x,y)=\frac{c}{a+b-c}\in [0,1]$.

Given splits Train, Val, and Test, we form similarity pairs. On datasets with large unique molecules, uniform random sampling is taken:
\begin{align}
Pair_{Train-Train} &= \{ T(x_i,x_j) : x_i,x_j\in Train,\, i\neq j \} \\
Pair_{Train-Val} &= \{ T(x_i,y_j) : x_i\in Train,\, y_j\in Val \} \\
Pair_{Train-Test} &= \{ T(x_i,x_j) : x_i\in Train,\, y_j\in Test \}
\end{align}

To test whether Train-Val distributions and Train-Test distributions differ, the two-sample Kolmogorov–Smirnov (KS) statistic is defined as:
\begin{equation}
D_{n,m} \;=\; \sup_{t\in\mathbb{R}} \bigl|\widehat{F}_n(t) - \widehat{G}_m(t)\bigr|,
\end{equation}
where $\widehat{F}_n$ and $\widehat{G}_m$ are the empirical CDFs. Under the null hypothesis $H_0$, the corresponding $p$-value has the Kolmogorov tail with the large-$Z$ asymptotic approximation ($Z = D_{n,m}\sqrt{n_{\mathrm{eff}}}$, where $n_{\mathrm{eff}} = \frac{nm}{n+m}$):
\begin{equation}
p \approx 2e^{-2Z^2} \quad (\text{for large } Z).
\end{equation}

For scaffold overlap using Murcko scaffolds $\sigma(\cdot)$, let $\Sigma(\mathcal{A})=\{\sigma(x):x\in\mathcal{A}\}$. The overlaps are defined as:
\begin{equation}
\mathrm{overlap}_{\mathrm{test\ in\ train}} = \frac{|\Sigma(\mathcal{A})\cap \Sigma(\mathcal{B})|}{|\Sigma(\mathcal{B})|},
\quad
\mathrm{Jaccard\ overlap} = \frac{|\Sigma(\mathcal{A})\cap \Sigma(\mathcal{B})|}{|\Sigma(\mathcal{A})\cup \Sigma(\mathcal{B})|}.
\end{equation}

\subsubsection{Spectral Entropy Analysis}
To assess distributional shifts from the spectral perspective, we employ spectral entropy \cite{li2021spectral}. For a spectrum with $n$ peaks of intensities $\{I_i\}_{i=1}^n$, the spectral entropy is defined as:
\begin{equation}
H = -\sum_{i=1}^{n} p_i \ln(p_i), \quad \text{where} \quad p_i = \frac{I_i}{\sum_{j=1}^{n} I_j}
\end{equation}
We compute spectral entropy for all spectra and apply the two-sample KS test to compare distributions.

\subsubsection{Critical Difference Diagram with Wilcoxon-Holm Method}
To evaluate the relative performance of $k$ models across $N$ datasets, the average rank $\bar{R}_j$ for model $C_j$ is calculated as:
\begin{equation}
    \bar{R}_j = \frac{1}{N} \sum_{i=1}^{N} r_{i,j}
\end{equation}
Statistical significance is determined using a Friedman test followed by a post-hoc Wilcoxon signed-rank test. To control the Family-Wise Error Rate across $m$ pairwise comparisons, we apply the Holm-Bonferroni correction. For the ordered $p$-values, a hypothesis $H_i$ is rejected when
\begin{equation}
    p_i \le \frac{\alpha}{m - i + 1} \quad (\alpha=0.05).
\end{equation}

\subsection{Detailed Featurizer Architectures}

\paragraph{Traditional Featurizers} 
We implemented a suite of traditional methods that convert molecular structures into fixed-length vector representations, primarily leveraging RDKit \cite{bento2020open} to encode structural and property-based features. We utilize \textbf{Morgan fingerprints} (generating bit vectors through circular hashing of atomic environments with specified radii and bit lengths) \cite{morgan1965generation}, \textbf{MACCS keys} (166 predefined structural fragments for binary encoding) \cite{durant2002reoptimization}, and \textbf{Physicochemical descriptors} (computing molecular properties such as exact molecular weight, logP, hydrogen bond donors and acceptors, rotatable bonds, topological polar surface area, formal charge, and ring counts). We also compute \textbf{Atom pair fingerprints} (encoding topological distances between atom pairs), \textbf{Extended-reduced Graph (ErG) fingerprints} (representing simplified molecular graphs) \cite{stiefl2006erg}, \textbf{Daylight fingerprints} (using path-based hashing to produce 2048-bit vectors) \cite{james2004daylight}, and \textbf{PubChem fingerprints} (881-bit vectors based on predefined substructures) \cite{helal2016public}. Finally, a concatenated descriptor featurizer combines Morgan, Morgan counts, MACCS, physicochemical, atom pair, and ErG features into a unified, high-dimensional representation.

\paragraph{Sequence-Based Featurizers}
Inspired by sequence processing in natural language models, we introduced featurizers that treat SMILES strings as sequential data. \textbf{One-hot encoding} transforms SMILES characters into categorical vectors, padding or truncating to a maximum sequence length with a predefined alphabet of 58 symbols (including atoms, bonds, and special tokens). \textbf{Explainable Substructure Partition fingerprint (ESPF)} \cite{huang2019explainable} employs byte-pair encoding (BPE) on SMILES strings using a ChEMBL-derived vocabulary, producing tokenized sequences with optional masking and length constraints for transformer compatibility. \textbf{ChemBERTa} \cite{chithrananda2020chemberta} utilizes the output embeddings generated by a pretrained masked language model applied to these tokenized sequences, obtaining fixed-dimensional molecular representations by post-processing the token embeddings.

\paragraph{Graph-Based Featurizers}
To exploit the inherent topological structure of molecules, we incorporated graph neural network-compatible featurizers using DGL and RDKit \cite{wang2019deep, bento2020open}. \textbf{Canonical featurizers} employ atom and bond encoders (atomic numbers, chirality, hybridization, bond types) with self-loops and optional virtual nodes for padding to a fixed node count. \textbf{AttentiveFP} \cite{xiong2019pushing} extends this framework with attention-based atom and bond features, capturing local environments through multi-head mechanisms. \textbf{3D featurizers} generate complete graphs from single conformers, embedding 3D coordinates and spatial distances using Alchemy-inspired node and edge features after conformer optimization via UFF force fields. Additionally, simplified molecular graphs convert molecules into PyTorch Geometric data objects encoded with sparse adjacency formats \cite{fey2019fast}.

\subsection{MetaData Embedding}
In our model architecture, metadata associated with mass spectra includes absolute collision energy (ACE), normalized collision energy (NCE), instrument type, precursor type, ion mode, and precursor $\mathbf{m/z}$. 

For ACE and NCE, binary indicators are created to flag missing values. The non-missing values are standardized by subtracting the mean and dividing by the standard deviation with missing values filled as -1 post-normalization. Categorical features are converted to one-hot encoding. If the number of unique categories in the current dataset differs from a predefined count, the one-hot vectors are zeroed out to maintain dimensionality consistency. Precursor $\mathbf{m/z}$ is binned into integer values and treated as a discrete feature. The final metadata embedding is formed by concatenating these processed components.

\subsection{Model Architectures}
\label{sec:model_architectures}
FlexMS adopts a modular design philosophy, decoupling the mass spectrum prediction pipeline into two distinct components: the Molecule Embedding Model (Embedder) and the Spectrum Predictor (Predictor). This modularity allows for the systematic benchmarking of arbitrary combinations of molecular encoders and prediction heads \cite{liebal2020machine}.

\subsubsection{Embedders}
The role of the embedder is to map the raw molecular features into a fixed-dimensional latent vector ($h_{mol}$), encapsulating essential structural and chemical information. FlexMS implements a robust suite of embedders covering different inductive biases:
\begin{itemize}
    \item \textbf{MLP-based Embedders:} Multi-Layer Perceptrons (MLPs) with residual connections map high-dimensional sparse vectors (e.g., Morgan, MACCS) into dense latent representations. Daylight, ESPF, and AttentiveFP fingerprints are trained via these networks, alongside an "All-Descriptor" concatenation.
    \item \textbf{Sequence-based Embedders:} We utilize 1D-CNNs to capture local sequence patterns from one-hot encodings \cite{hirohara2018convolutional}. Furthermore, we incorporate Transformer-based encoders (including DeepPurpose \cite{huang2020deeppurpose}, ChemBERTa \cite{chithrananda2020chemberta}, and MoleBERT \cite{xia2023mole}) to capture long-range chemical dependencies through self-attention mechanisms.
    \item \textbf{Graph Neural Networks (GNNs):} We implement standard message-passing architectures for graph inputs, including GCN \cite{kipf2016semi}, GAT \cite{velivckovic2017graph}, and GIN \cite{xu2018powerful}. To evaluate advanced spatial biases, we also include AttentiveFP \cite{xiong2019pushing} and GraphormerV2 (GFv2) \cite{yang2021graphformers}, utilizing sophisticated spatial encodings.
\end{itemize}

\subsubsection{Predictors}
The predictor serves as the decoding module, taking the concatenated vector of the molecular embedding ($h_{mol}$) and the metadata embedding ($h_{meta}$) as input to generate the final binned mass spectrum. We adapt three distinct predictor architectures representing different modeling paradigms:
\begin{itemize}
    \item \textbf{MassFormer Predictor:} A high-capacity MLP design featuring a bidirectional prediction mechanism \cite{young2024tandem}. It explicitly models two simultaneous processes: the forward generation of fragments and the reverse neutral loss from the precursor ion. A learnable gating mechanism dynamically weighs the contribution of these two streams, optimizing peak capture across both low and high $m/z$ regions.
    \item \textbf{NEIMS Predictor:} A streamlined, efficiency-focused MLP architecture that prioritizes inference speed \cite{wei2019rapid}. While mathematically simpler than MassFormer, it retains the bidirectional logic required to handle the physical constraints of fragmentation, providing a highly efficient and robust baseline.
    \item \textbf{MolMS Predictor:} An advanced encoder-decoder structure emphasizing deep residual learning \cite{hong20233dmolms}. The decoding block utilizes multiple fully connected layers integrated with Residual Blocks. This prevents vanishing gradient issues in deeper networks, theoretically enabling the model to capture highly complex, non-linear relationships between molecular structures and spectral intensity outputs.
\end{itemize}

\subsection{Evaluation Metrics Formulation}
\label{sec:evaluation_metric}

\subsubsection{Generative Prediction Metrics}
\begin{itemize}
    \item \textbf{Cosine Similarity:} Let $\mathbf{y}_{\mathrm{pred}}, \mathbf{y}_{\mathrm{true}} \in \mathbb{R}_{\ge 0}^{B}$ denote the predicted and observed binned spectra over $B$ mass bins. Cosine similarity is defined as
    \begin{equation}
        \mathrm{CosSim}(\mathbf{y}_{\mathrm{pred}}, \mathbf{y}_{\mathrm{true}})
        =
        \frac{\mathbf{y}_{\mathrm{pred}}^\top \mathbf{y}_{\mathrm{true}}}
        {\|\mathbf{y}_{\mathrm{pred}}\|_2 \, \|\mathbf{y}_{\mathrm{true}}\|_2}.
    \end{equation}

    \item \textbf{Jensen--Shannon Similarity:} For distribution-level comparison, the normalized spectra are first defined as
    \begin{equation}
        \tilde{\mathbf{y}}_{\mathrm{pred}} = \frac{\mathbf{y}_{\mathrm{pred}}}{\sum_{b=1}^{B} y_{\mathrm{pred},b}},
        \qquad
        \tilde{\mathbf{y}}_{\mathrm{true}} = \frac{\mathbf{y}_{\mathrm{true}}}{\sum_{b=1}^{B} y_{\mathrm{true},b}},
    \end{equation}
    with mixture distribution
    \begin{equation}
        \mathbf{m} = \frac{1}{2}\left(\tilde{\mathbf{y}}_{\mathrm{pred}} + \tilde{\mathbf{y}}_{\mathrm{true}}\right).
    \end{equation}
    The Jensen--Shannon similarity is then computed as
    \begin{equation}
        \mathrm{JSSim}(\mathbf{y}_{\mathrm{pred}}, \mathbf{y}_{\mathrm{true}})
        =
        1 - \frac{1}{2}
        \left[
            D_{\mathrm{KL}}(\tilde{\mathbf{y}}_{\mathrm{pred}} \,\|\, \mathbf{m})
            +
            D_{\mathrm{KL}}(\tilde{\mathbf{y}}_{\mathrm{true}} \,\|\, \mathbf{m})
        \right].
    \end{equation}

    \item \textbf{Spectral Coverage:} Spectral coverage measures peak-level recall after thresholding with $\tau$. Let
    \begin{equation}
        \mathbf{1}_{\tau}(y_b) = \mathbf{1}[y_b > \tau].
    \end{equation}
    Coverage is defined as
    \begin{equation}
        \mathrm{Coverage}(\mathbf{y}_{\mathrm{pred}}, \mathbf{y}_{\mathrm{true}})
        =
        \frac{\sum_{b=1}^{B} \mathbf{1}_{\tau}(y_{\mathrm{pred},b}) \, \mathbf{1}_{\tau}(y_{\mathrm{true},b})}
        {\sum_{b=1}^{B} \mathbf{1}_{\tau}(y_{\mathrm{true},b})}.
    \end{equation}
\end{itemize}
\subsubsection{Retrieval Metrics}
\begin{itemize}
    \item \textbf{Raw Rank:} Let $\mathcal{Q}$ denote the set of query spectra. For each query $q \in \mathcal{Q}$, let $C_q$ be the associated candidate set, let $c_q^\star \in C_q$ denote the ground-truth molecule, and let
    \begin{equation}
        s_q(c) = \mathrm{CosSim}\!\left(\hat{\mathbf{y}}_{q,c}, \mathbf{y}_q\right)
    \end{equation}
    be the cosine similarity between the predicted spectrum for candidate $c$ under the metadata of query $q$ and the observed query spectrum $\mathbf{y}_q$. The raw rank of the ground-truth molecule is then defined as
    \begin{equation}
        \mathrm{Rank}(q) = 1 + \sum_{c \in C_q \setminus \{c_q^\star\}} \mathbf{1}\!\left[s_q(c) > s_q(c_q^\star)\right].
    \end{equation}
    Lower values indicate better retrieval performance, and $\mathrm{Rank}(q)=1$ corresponds to a correct top-ranked identification.

    \item \textbf{Normalized Rank:} To account for varying candidate-pool sizes, the normalized rank is defined as
    \begin{equation}
        \mathrm{NormRank}(q) = \frac{\mathrm{Rank}(q)}{|C_q|}.
    \end{equation}
    Aggregate retrieval performance is summarized by the query-averaged rank and normalized rank:
    \begin{equation}
        \overline{\mathrm{Rank}} = \frac{1}{|\mathcal{Q}|}\sum_{q \in \mathcal{Q}} \mathrm{Rank}(q),
        \qquad
        \overline{\mathrm{NormRank}} = \frac{1}{|\mathcal{Q}|}\sum_{q \in \mathcal{Q}} \mathrm{NormRank}(q).
    \end{equation}

    \item \textbf{Top-$K$ Accuracy:} Top-$K$ accuracy measures whether the ground-truth molecule appears within the first $K$ ranked candidates:
    \begin{equation}
        \mathrm{Top}\mbox{-}K = \frac{1}{|\mathcal{Q}|}\sum_{q \in \mathcal{Q}} \mathbf{1}\!\left[\mathrm{Rank}(q) \le K\right].
    \end{equation}
    In the experiments, $K \in \{1,5,10\}$ is reported.

    \item \textbf{Top-$p\%$ Accuracy:} Top-$p\%$ accuracy measures whether the ground-truth molecule lies within the top $p\%$ of the candidate pool:
    \begin{equation}
        \mathrm{Top}\mbox{-}p\% = \frac{1}{|\mathcal{Q}|}\sum_{q \in \mathcal{Q}} \mathbf{1}\!\left[\mathrm{NormRank}(q) \le p/100\right].
    \end{equation}
    Results are reported for $p \in \{1,5,10\}$.
\end{itemize}

\subsection{Hyperparameters and Training Details}
\label{app:hyperparameters}
We fixed specific hyperparameters across the benchmark:
\begin{itemize}
    \item \textbf{CNN:} 3 1D-Conv blocks with channels 32, 64, 96, kernel size 4, and an output dimension of 256.
    \item \textbf{GAT:} 3 DGL GAT blocks with ReLU, hidden dimension 64, and output dimension 64.
    \item \textbf{GCN:} 1 DGL GCN block with ReLU, 3 hidden layers of dimension 64, and output dimension 256.
    \item \textbf{GIN:} 4 DGL GINConv layers with batch normalization and ReLU. Each layer contains a 2-layer MLP (hidden dimension 64). Output dimension is 64.
    \item \textbf{Transformer:} 8 transformer layers, 8 attention heads with hidden size 512, and output embedding dimension of 128.
    \item \textbf{AttentiveFP:} 2 AttentiveFPGNN layers with node feature size 39, edge feature size 11, and output dimension 64.
    \item \textbf{MLPs (ESPF, Daylight, All-descriptors):} Architecture features 3 hidden layers of sizes 1024, 256, and 64 with ReLU activations.
\end{itemize}

\paragraph{Experimental and Training Configurations}
To ensure reproducibility and fair evaluation, most critical training details were locked across dataset variants unless specified otherwise:
\begin{itemize}
    \item \textbf{Optimizer and Scheduler:} All models are optimized using the \textbf{AdamW} optimizer at a base learning rate of $10^{-4}$ (unless mentioned) and $\text{weight\_decay}=0.0$.
    \item \textbf{Loss Function:} Training explicitly optimizes a modified Cosine Loss defined primarily as $(1 - \mathrm{mean}(\text{cosine\_sim}))$.
    \item \textbf{Epochs and Early Stopping:} We train all regressors for a maximum of \textbf{100 epochs}. Early stopping evaluates the validation cosine score, applying a strict early-stopping, reverting to the checkpoint with the highest validation metric.
    \item \textbf{Batch Size:} The standard training batch size enforced for main experimental suite is \textbf{512}.
    \item \textbf{Random Seeds:} Datasets are seeded deterministically (seed 42) to isolate architectural impact reliably. 
    \item \textbf{GFv2 / Foundation Details:} Models leveraging massive-scale pretrianing (such as GraphormerV2) were \textbf{fully finetuned} end-to-end to aggressively adapt their chemical spatial awareness directly toward the MS prediction manifold rather than keeping attention weights frozen.
\end{itemize}

\section{Additional Experimental Results and Figures}
\label{app:additional_results}

This section provides additional empirical results and comprehensive visualizations that complement the core findings presented in the main manuscript, including detailed data ablation studies, learning rate sensitivity analyses, and the extended effects of molecular pretraining.

\subsection{Comprehensive analysis on predictor and embedder}
\begin{figure}[htbp]
    \centering
    \includegraphics[width=9cm]{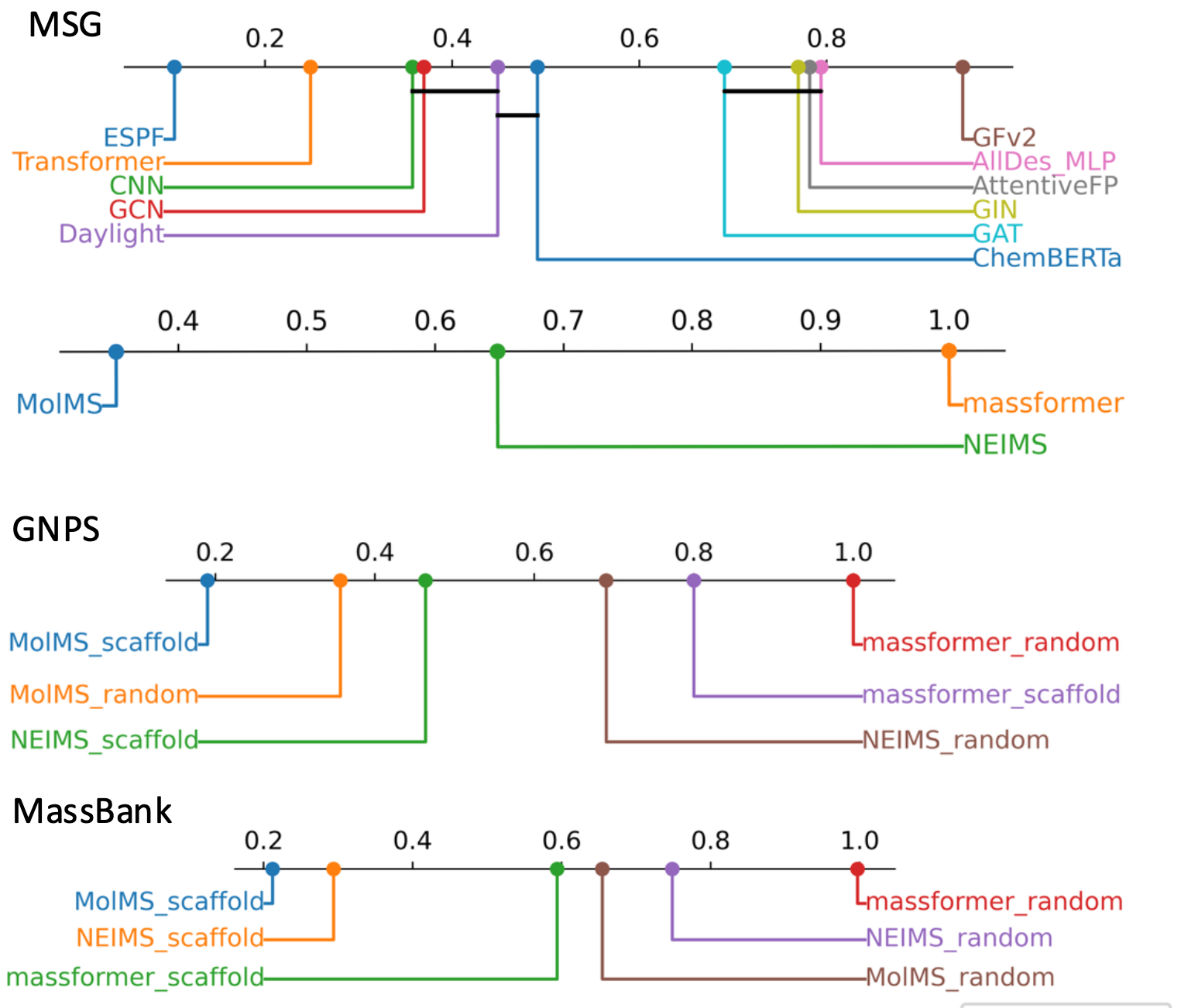}
    \caption{Critical difference diagrams of different embedders and predictors on the MassSpecGym, GNPS and MassBank datasets, obtained by combining results across embedder-predictor pairs, resolutions and datasets, and applying the Wilcoxon-Holm test to detect pairwise significance.}
    \label{fig:benchmark}
\end{figure}

Figure~\ref{fig:benchmark} shows the CDF about predictor and embedder. It illustrates the statistical performance hierarchy of the evaluated embedders and predictors, where positions further to the right signify superior predictive accuracy. In the MSG dataset, graph-based embedders such as GFv2, AttentiveFP, and GIN demonstrate significantly higher efficacy compared to sequence-based models. Furthermore, massformer consistently emerges as the top-ranking predictor across all three datasets—MSG, GNPS, and MassBank—maintaining a statistically significant lead over NEIMS and MolMS in both random and scaffold split scenarios.

\subsection{Retrieval benchmark performance on CASMI22}

The comparison between CASMI 2016 and CASMI 2022 reveals that performance degradation is primarily driven by the substantially larger candidate pool in the 2022. This harder retrieval regime amplifies the difficulty of placing the correct structure near the top of the ranking: a model that achieves Recall@10 of 60\% on CASMI 2016 may drop to under 19\% on CASMI 2022, not necessarily because its underlying representations have deteriorated, but because the prior probability of correct identification decreases proportionally with pool size. The candidate-count distributions further confirm this asymmetry—CASMI 2016 exhibits a long-tailed distribution with many small pools, while CASMI 2022 is concentrated near the upper cap, indicating that most queries face a similarly challenging high-prior regime.

Despite the absolute performance drop, the relative ordering of embedders remains broadly consistent across both benchmarks. GFv2 maintains the strongest embedder, and graph-based methods continue to outperform. This pattern suggests that pretrained graph representations capture structural features that are both necessary for spectrum reconstruction and robust to variations in retrieval difficulty. However, the magnitude of GFv2's advantage narrows considerably in CASMI 2022, particularly for top-ranked candidates (Recall@1), indicating that even the best available molecular representations face fundamental limits when discriminating among chemically plausible candidates in extremely large search spaces.

\begin{table}[ht]
\centering
\caption{Retrieval benchmark performance evaluated on the CASMI 2022 dataset using different predictors. Metrics include Recall@1 (R@1), Recall@10 (R@10), and Normalized Rank (NormR $\downarrow$) across candidate pools. Best, second best, and third best results in each column are denoted by \textbf{bold}, \underline{\underline{doubleline}}, \underline{underline}, respectively.}
\label{tab:retrieval_metrics_casmi22}
\resizebox{\textwidth}{!}{
\begin{tabular}{lccccccccc}
\toprule
 & \multicolumn{3}{c}{\textbf{MassFormer}} & \multicolumn{3}{c}{\textbf{NEIMS}} & \multicolumn{3}{c}{\textbf{MolMS}} \\
\cmidrule(lr){2-4} \cmidrule(lr){5-7} \cmidrule(lr){8-10}
& \textbf{R@1} & \textbf{R@10} & \textbf{NormR$\downarrow$} & \textbf{R@1} & \textbf{R@10} & \textbf{NormR$\downarrow$} & \textbf{R@1} & \textbf{R@10} & \textbf{NormR$\downarrow$} \\
\midrule
CNN          & 0.0044 & 0.0306 & 0.3332 & 0.0015 & 0.0175 & 0.3576 & 0.0000 & 0.0087 & 0.3906 \\
GCN          & 0.0044 & 0.0830 & 0.2043 & 0.0131 & 0.0859 & 0.2184 & 0.0087 & 0.0611 & 0.2624 \\
GIN          & 0.0131 & 0.0786 & 0.2109 & 0.0073 & 0.0830 & 0.2215 & 0.0044 & 0.0699 & 0.2472 \\
GAT          & 0.0087 & 0.0480 & 0.2473 & 0.0044 & 0.0393 & 0.2378 & 0.0044 & 0.0262 & 0.2510 \\
Trans        & 0.0087 & 0.0262 & 0.3863 & 0.0029 & 0.0116 & 0.3821 & 0.0000 & 0.0000 & 0.3967 \\
ESPF         & 0.0044 & 0.0131 & 0.3844 & 0.0029 & 0.0116 & 0.3689 & 0.0044 & 0.0131 & 0.3557 \\
DL           & 0.0087 & 0.0917 & 0.2046 & 0.0058 & 0.0902 & 0.2128 & 0.0000 & 0.0786 & 0.2348 \\
AttnFP       & 0.0044 & 0.0786 & 0.1918 & 0.0044 & 0.0655 & 0.2225 & 0.0000 & 0.0393 & 0.2852 \\
AllDes       & 0.0087 & 0.0917 & 0.2046 & 0.0058 & 0.0902 & 0.2128 & 0.0000 & 0.0786 & 0.2348 \\
GFv2         & \underline{0.0218} & \textbf{0.1878} & \textbf{0.1761} & \textbf{0.0277} & \textbf{0.1572} & \textbf{0.1767} & \underline{0.0087} & \textbf{0.1004} & \textbf{0.2017} \\
\bottomrule
\end{tabular}
}
\end{table}

\subsection{Cross-Domain Transfer Learning Analysis (Complete)}
\label{app:cross_domain}
\begin{figure}[htbp]
\centering
\includegraphics[width=\textwidth]{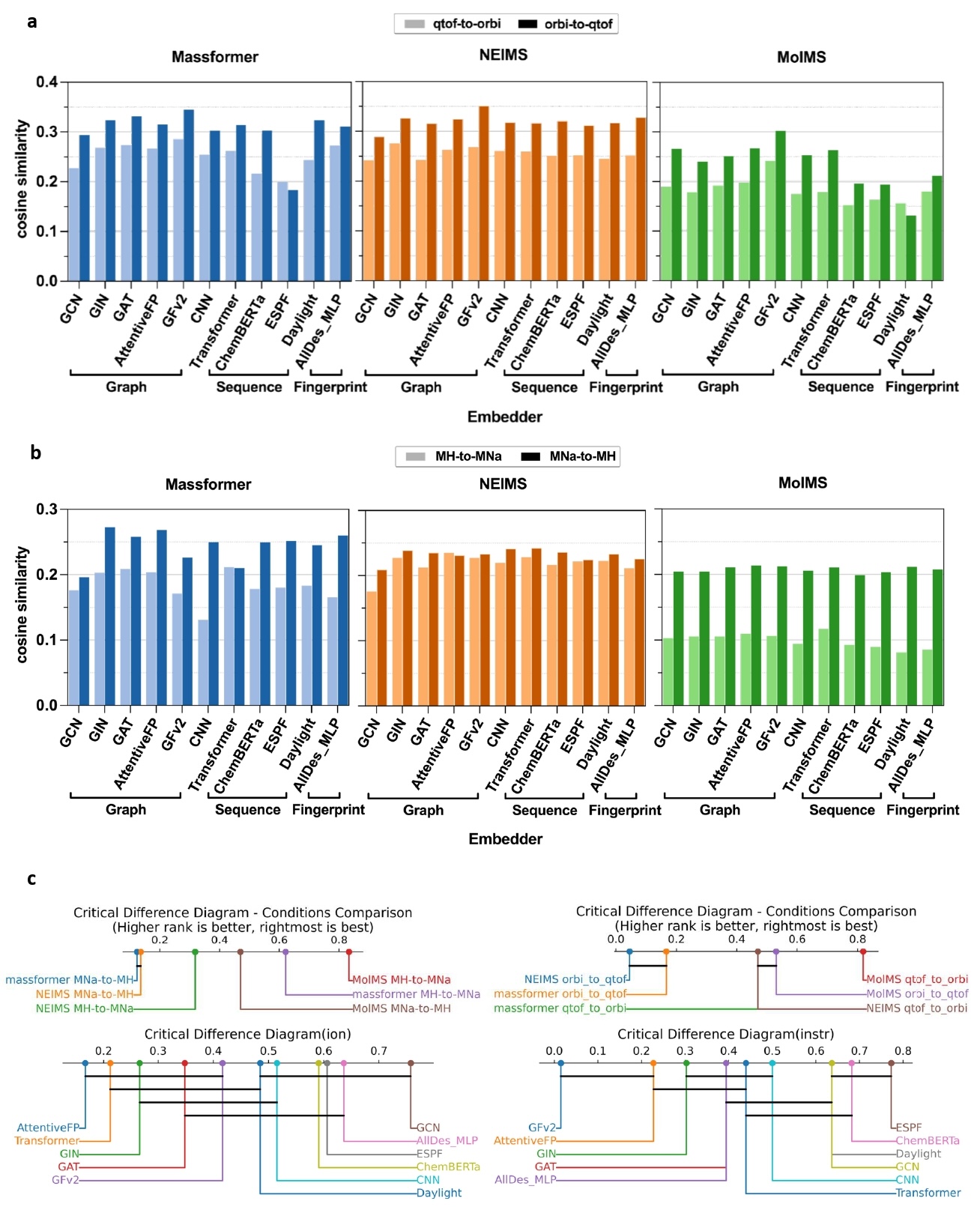}
\caption{\textbf{(a)(b)} Detailed performance comparison of domain transfer, ionization mode transfer and instrument type transfer. \textbf{(c)}The critical difference diagram of different transfer case and embedders. Left is ion-transfer case and right is instrument-transfer case.}
\label{fig:transfer_results}
\end{figure}

Figure~\ref{fig:transfer_results} shows the detailed results about retrieval analysis.

\subsection{Same-Domain Simulation Analysis}
\label{app:same_domain}
To complement the cross-domain transfer learning analysis, we also evaluated within-domain predictive performance. Table~\ref{tab:transfer_same} demonstrates the spectrum reconstruction fidelity (measured by cosine similarity) when models are trained and tested on the same instrument type (Orbitrap, QTOF) or the same ionization mode ([M+H]$^+$, [M+Na]$^+$). 

Consistently, the pretrained graph model GFv2 outperforms almost all other representations across different predictors and sub-domains, confirming that the structural priors captured during pretraining are highly beneficial for capturing within-domain fragmentation rules. Furthermore, prediction accuracy on QTOF spectra generally surpasses Orbitrap spectra across most architectures. In terms of ionization modes, [M+H]$^+$ adducts systematically yield higher reconstruction fidelity than [M+Na]$^+$ adducts. The notable drop in performance for [M+Na]$^+$ (especially with the MolMS predictor) suggests that sodium-adduct fragmentation pathways are either underrepresented in the training data or inherently more difficult to simulate due to their distinct rearrangement mechanisms.

\begin{table}[htbp]
\centering
\caption{Within-domain transfer learning performance (cosine similarity, number=10). Best results per row are in bold. Transfer directions (Train $\to$ Test) are denoted using: Q (QTOF) and O (Orbitrap) for instrument types; H ([M+H]$^+$) and Na ([M+Na]$^+$) for ionization modes.}
\label{tab:transfer_same}
\resizebox{\textwidth}{!}{
\begin{tabular}{llc*{11}{>{\centering\arraybackslash}p{0.8cm}}}
\toprule
\textbf{Task} & \textbf{Pred} & \textbf{Subset} & \textbf{CNN} & \textbf{GCN} & \textbf{GIN} & \textbf{GAT} & \textbf{Trans} & \textbf{ESPF} & \textbf{DL} & \textbf{AttnFP} & \textbf{AllDes} & \textbf{CBT} & \textbf{GFv2} \\
\midrule
\multirow{6}{*}{Inst.}
& \multirow{2}{*}{Mass} & Orbi & 0.2862 & 0.2963 & 0.2972 & 0.2898 & 0.2836 & 0.2807 & 0.2915 & 0.2975 & 0.3121 & 0.2975 & \textbf{0.3392} \\
& & QTOF & 0.3233 & 0.3220 & 0.3493 & 0.3357 & 0.3232 & 0.3254 & 0.3583 & 0.3512 & 0.3670 & 0.3223 & \textbf{0.3883} \\
\cmidrule{3-14}
& \multirow{2}{*}{NEIMS} & Orbi & 0.2797 & 0.2909 & 0.3096 & 0.2885 & 0.2618 & 0.2843 & 0.2953 & 0.3047 & 0.3073 & 0.2880 & \textbf{0.3377} \\
& & QTOF & 0.3071 & 0.3061 & 0.3486 & 0.3264 & 0.3263 & 0.3229 & 0.3390 & 0.3457 & 0.3497 & 0.3348 & \textbf{0.3690} \\
\cmidrule{3-14}
& \multirow{2}{*}{MolMS} & Orbi & 0.2083 & 0.2190 & 0.2270 & 0.2233 & 0.2030 & 0.1949 & 0.2088 & 0.2233 & 0.2092 & 0.2162 & \textbf{0.2514} \\
&  & QTOF & 0.2714 & 0.2811 & 0.2992 & 0.2910 & 0.2726 & 0.2606 & 0.2924 & 0.3043 & 0.2999 & 0.2953 & \textbf{0.3148} \\

\midrule
\multirow{6}{*}{Ion.}
& \multirow{2}{*}{Mass} & MH & 0.3129 & 0.3282 & 0.3271 & 0.3270 & 0.3139 & 0.3011 & 0.3084 & 0.3371 & 0.3345 & 0.3278 & \textbf{0.3796} \\
& & Na & 0.2434 & 0.2180 & 0.2594 & 0.2366 & 0.2483 & 0.2533 & 0.2288 & 0.2402 & 0.2343 & 0.2408 & \textbf{0.2604} \\
\cmidrule{3-14}
& \multirow{2}{*}{NEIMS} & MH & 0.2934 & 0.2974 & 0.3321 & 0.3057 & 0.2997 & 0.2995 & 0.3159 & 0.3213 & 0.3274 & 0.3090 & \textbf{0.3718} \\
& & Na & 0.2760 & 0.2548 & 0.2607 & 0.2667 & 0.2427 & 0.2651 & 0.2438 & 0.2605 & 0.2614 & 0.2752 & \textbf{0.2824} \\
\cmidrule{3-14}
& \multirow{2}{*}{MolMS} & MH & 0.2447 & 0.2710 & 0.2791 & 0.2638 & 0.2489 & 0.2307 & 0.2561 & 0.2722 & 0.2681 & 0.2560 & \textbf{0.3099} \\
& & Na & 0.1229 & 0.1252 & 0.1295 & 0.1305 & 0.1302 & 0.1277 & \textbf{0.1410} & 0.1324 & 0.1246 & 0.1341 & 0.1321 \\

\bottomrule
\end{tabular}
}
\end{table}

\subsection{Learning Rate Sensitivity on Mass Prediction}
\label{learning_rate_pred}
\begin{figure}[htbp]
    \centering
    \includegraphics[width=\textwidth]{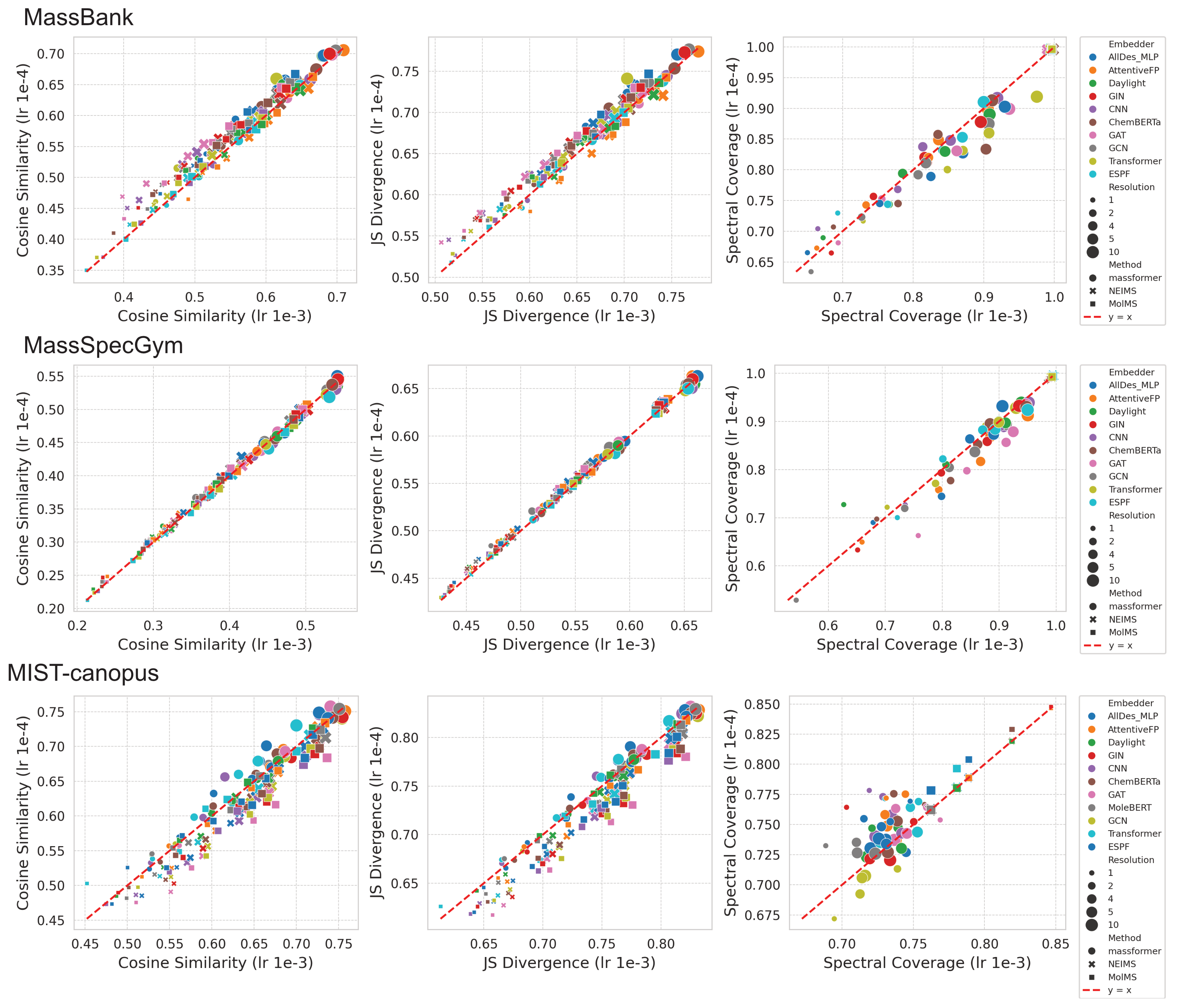}
    \caption{Cosine similarity, Jensen-Shannon similarity, and coverage across different learning rates and conditions. The analysis evaluates all combinations of embedders, predictors and resolutions across 3 datasets. Notably, the GNPS dataset is excluded from this specific ablation study due to its massive scale, which poses significant computational challenges and excessive training time.}
    \label{fig:FlexMS-LR-supp}
\end{figure}

Figure~\ref{fig:FlexMS-LR-supp} indicates that learning-rate effects are generally smaller than architecture effects, but they remain dataset-dependent. Most points lie close to the diagonal, which means that switching from $10^{-3}$ to $10^{-4}$ rarely changes the overall ranking of methods in a dramatic way. However, MassBank and MassSpecGym more often benefit from the smaller learning rate, consistent with their noisier and more heterogeneous optimization landscapes. This observation reinforces the main-text argument that hyperparameter selection in binned spectrum prediction should follow dataset characteristics rather than a single universal default.

\subsection{Learning Rate Sensitivity on Molecule Retrieval}
\label{app}
\begin{figure}[htbp]
    \centering
    \includegraphics[width=\textwidth]{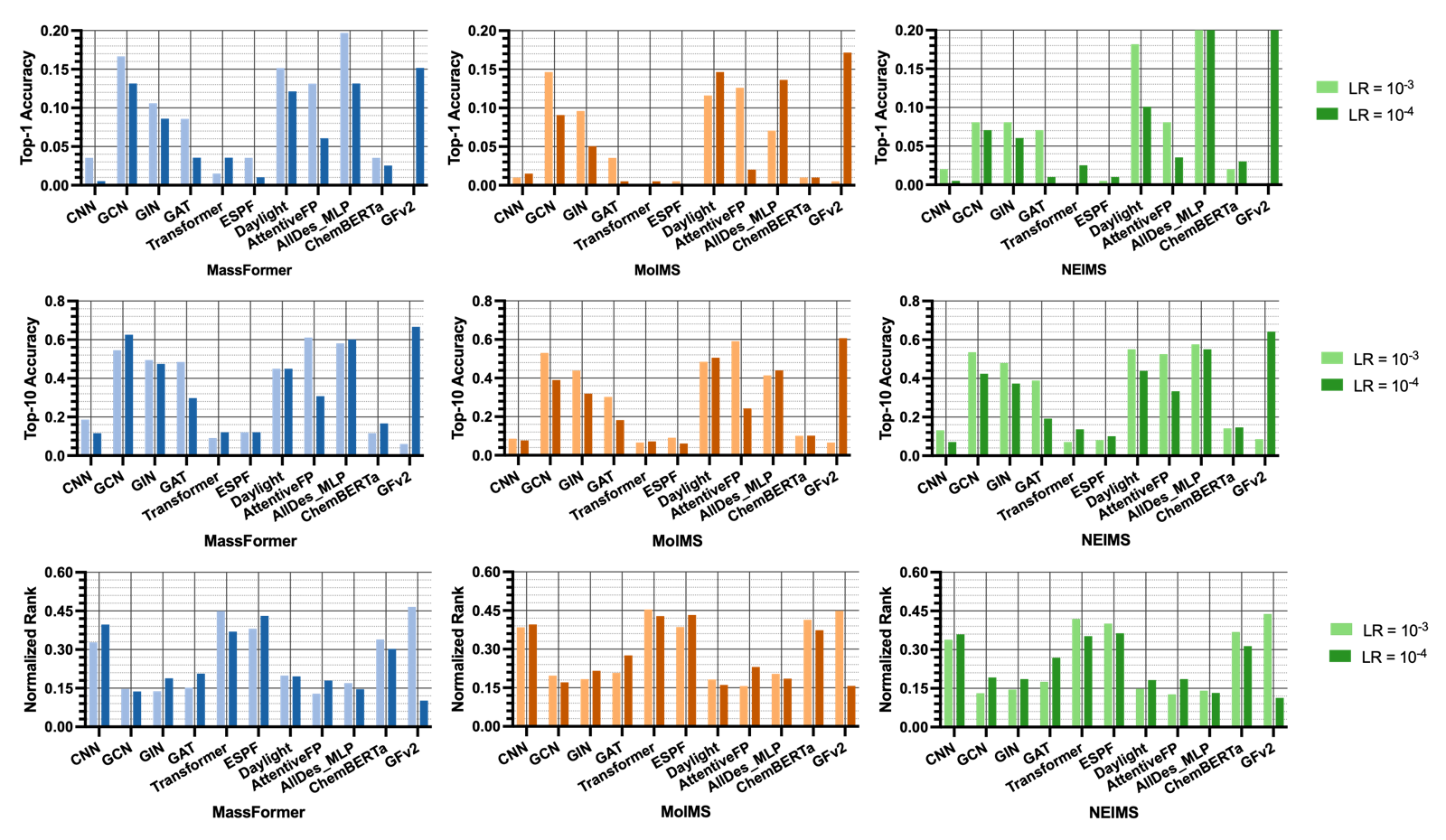}
    \caption{Retrieval performance metrics on the CASMI16 dataset across different learning rates. Various metrics were evaluated for this task to determine optimal convergence behaviors.}
    \label{fig:FlexMS-LR-retrieval}
\end{figure}

Figure~\ref{fig:FlexMS-LR-retrieval} extends the learning-rate comparison ($10^{-3}$ vs. $10^{-4}$) from direct spectrum reconstruction to downstream retrieval on the CASMI16 dataset. Based on our mass spectrum prediction evaluations, we established two key conclusions: (1) learning rate effects in generative modeling are generally secondary to architectural differences, and (2) optimal learning rates are typically dataset-dependent, with smoother optimization landscapes permitting $10^{-3}$ and noisier, heterogeneous spectra strictly requiring $10^{-4}$.

However, evaluating these models on downstream candidate retrieval reveals a markedly different sensitivity profile. While the performance gap between $10^{-3}$ and $10^{-4}$ remains relatively modest for classical fingerprint-based models (e.g., Daylight, ESPF) and 1D-CNNs, deep graph-based architectures---particularly pretrained models like GFv2---exhibit profound optimization instability. Specifically, across Massformer and MolMS predictors, GFv2 experiences a catastrophic collapse in Top-10 accuracy when fine-tuned at a higher learning rate of $10^{-3}$ (dropping to $\sim 0.06$), whereas reducing the learning rate to $10^{-4}$ restores performance to competitive levels ($\sim 0.60 - 0.66$). This stark contrast demonstrates that complex, high-capacity structural encoders are deeply susceptible to optimization divergence or catastrophic forgetting at higher learning rates. Consequently, downstream retrieval tasks heavily amplify structural instabilities that may appear benign under average generative metrics, emphasizing that retrieval-oriented model and hyperparameter selection cannot rely solely on reconstruction robustness.

\subsection{Effects of Molecular Pretraining (Complete)}
\label{sup:pretrain_full}
\begin{figure}[htbp]
    \centering
    \includegraphics[width=\textwidth]{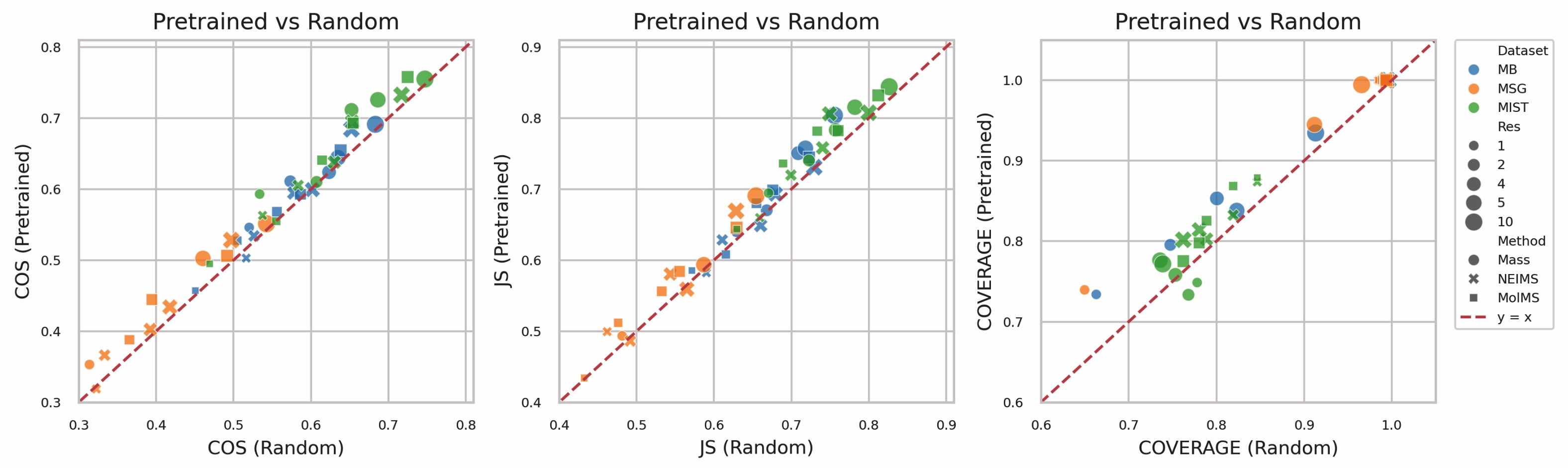}
    \caption{Performance metrics comparison between pretrained MoleBERT and random-initialized MoleBERT. Various metrics and datasets (MassBank, NPLIB1(MIST), and MassSpecGym) are evaluated to demonstrate the structural extrapolation capabilities acquired through self-supervised pretraining.}
    \label{fig:FlexMS-pretrained}
\end{figure}

Figure~\ref{fig:FlexMS-pretrained} shows that most pretrained-versus-random comparisons lie above the identity line, confirming that self-supervised molecular pretraining usually improves downstream spectrum prediction. The gains are more consistent for cosine similarity and Jensen-Shannon similarity than for spectral coverage, suggesting that pretraining mainly sharpens global spectral fidelity rather than merely recovering more peaks. The improvement is also observed across multiple datasets and predictor heads, which indicates that the benefit is not tied to a single benchmark configuration. Overall, these results support the interpretation that pretrained molecular representations transfer chemically meaningful regularities that remain useful in binned MS/MS prediction.

\subsection{Resolution Effects and Architectural Robustness in Mass Spectrum Simulation} 

\begin{figure}[htbp]
\centering
\includegraphics[width=\textwidth]{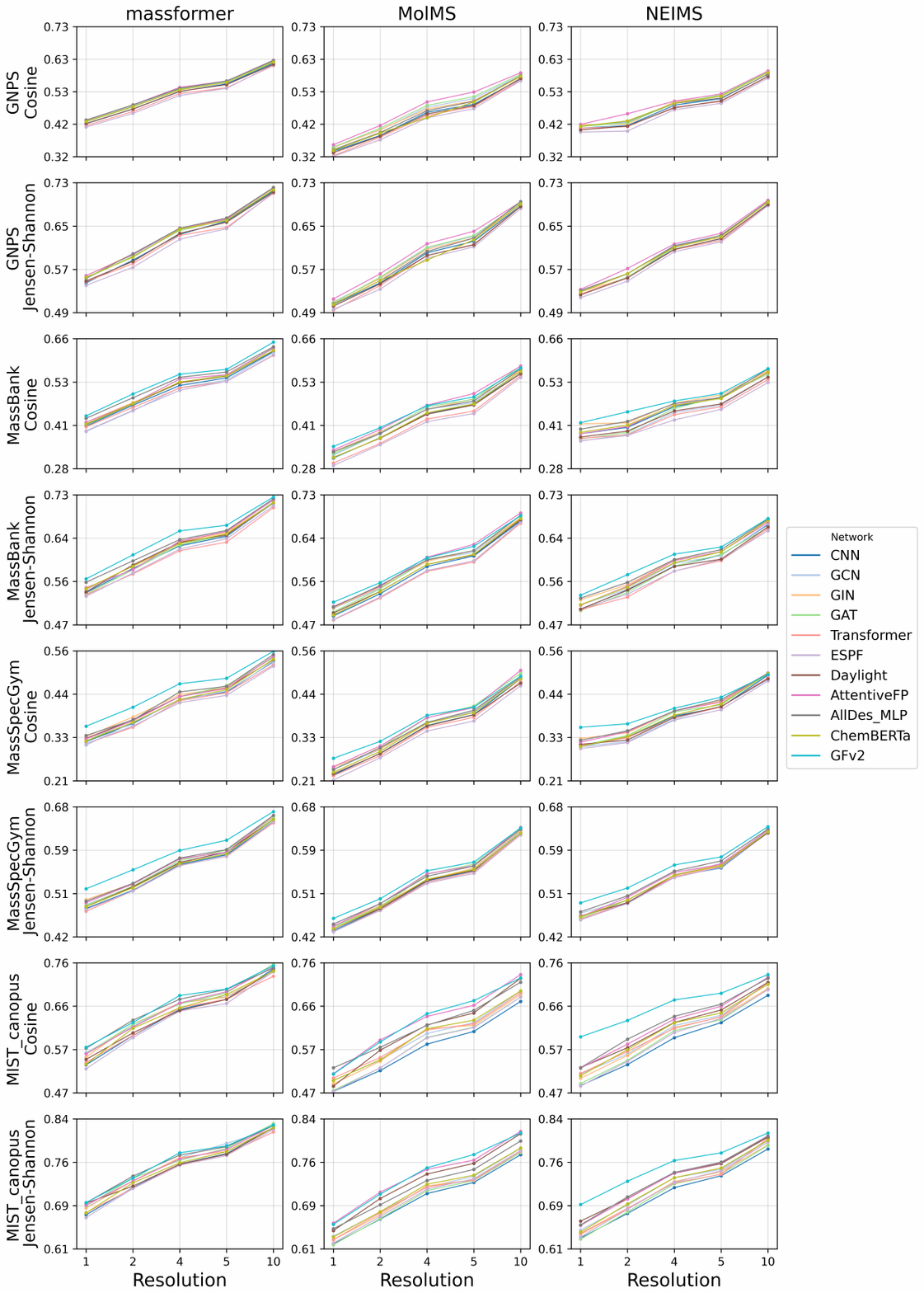}
\caption{Performance of different embedder-predictor combinations with different resolutions on GNPS dataset (random split), MassBank dataset (random split), MassSpecGym dataset and NPLIB1(MIST) dataset. }
\label{fig:resolution}
\end{figure}

Our systematic evaluation in Figure~\ref{fig:resolution} of spectral resolution (bin widths ranging from 1 Da to 10 Da) reveals a non-linear performance trajectory in the benchmark. This trend is primarily a mathematical effect of reduced output dimensionality rather than a true accuracy gain, it serves as a valuable structural test.

We identified a distinct architectural nuance in how models respond to resolution changes: graph-based embedders exhibit stable and monotonic improvements as resolution coarsens, indicating robust structural feature extraction. In contrast, sequence-based and pre-trained language representations exhibit more erratic performance trajectories, highlighting differential robustness to variations in spectral granularity.

\subsection{Resolution Effects and Architectural Robustness in Retrieval}
Consistent with the simulation findings, retrieval benchmarks on CASMI 2016 and CASMI 2022 exhibit a non-linear performance trajectory across resolution settings. Coarser resolutions universally yield improved retrieval accuracy, with the most substantial gains occurring between 1 Da and 2 Da, after which further aggregation provides diminishing returns. Graph-based embedders (GCN, GIN, GAT, GFv2) demonstrate stable, monotonic improvements in normalized rank as resolution coarsens, reflecting robust structural feature extraction invariant to spectral granularity.

Sequence-based and pre-trained representations (Transformer, ESPF, ChemBERTa) display erratic trajectories across both benchmarks, with non-monotonic fluctuations between resolution settings. This architectural divide is particularly evident in Top-5\% and Top-10\% accuracy trends: graph-based models maintain consistent ranking quality across resolutions, while sequence-based models exhibit sensitivity that can cause degradation at certain coarsening thresholds. The pattern suggests molecular graph representations remain stable under spectral aggregation, whereas SMILES-derived embeddings lose critical substructural information when peaks are binned, leading to inconsistent candidate ranking on real-world identification tasks.

\begin{figure}[htbp]
    \centering
    \includegraphics[width=\textwidth]{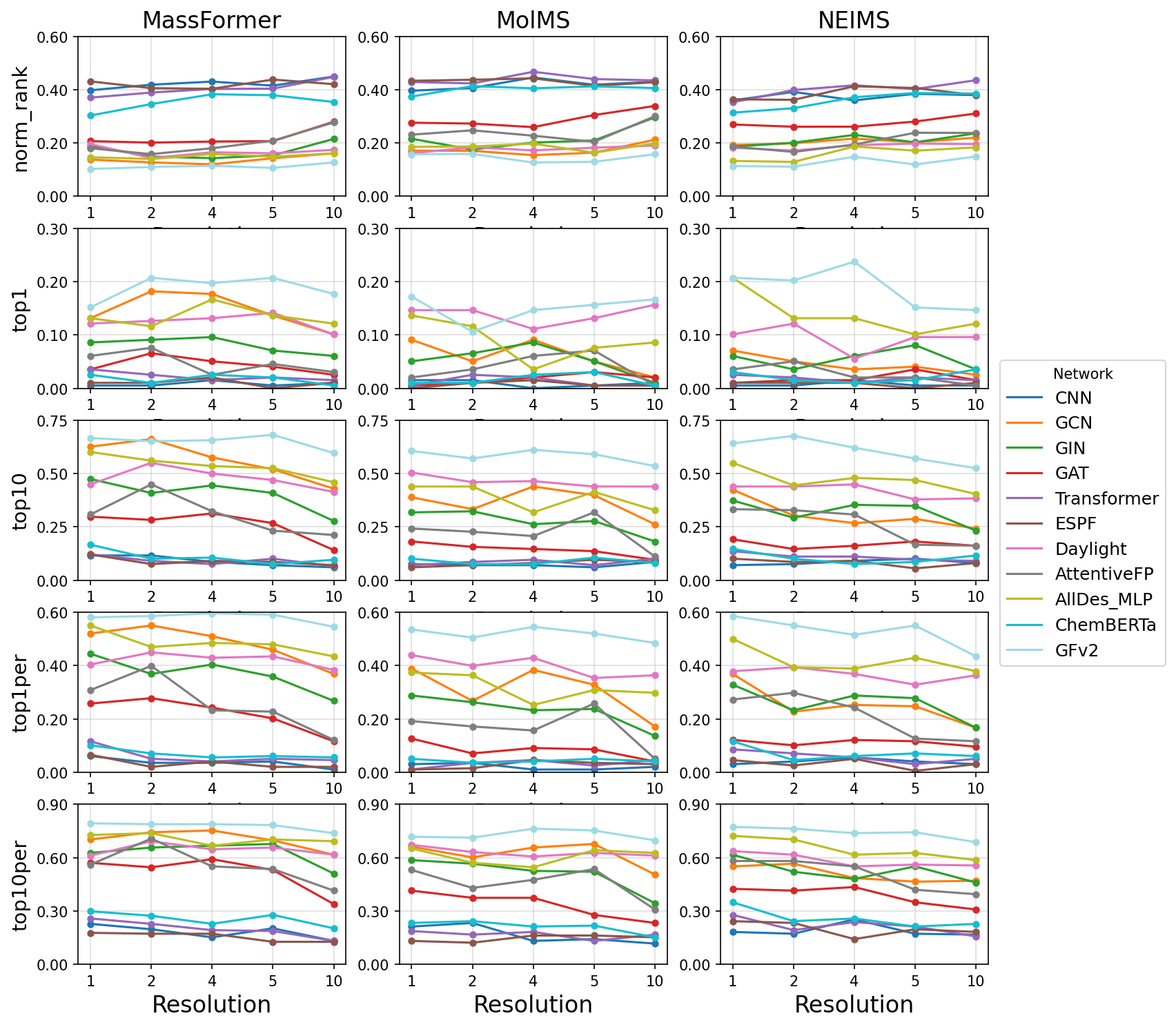}                                
    \caption{Performance of different embedder-predictor combinations with different resolutions on the CASMI 2016 retrieval benchmark.}
    \label{fig:retrieval-casmi2016}
\end{figure}                                                             

\begin{figure}[htbp]
    \centering
    \includegraphics[width=\textwidth]{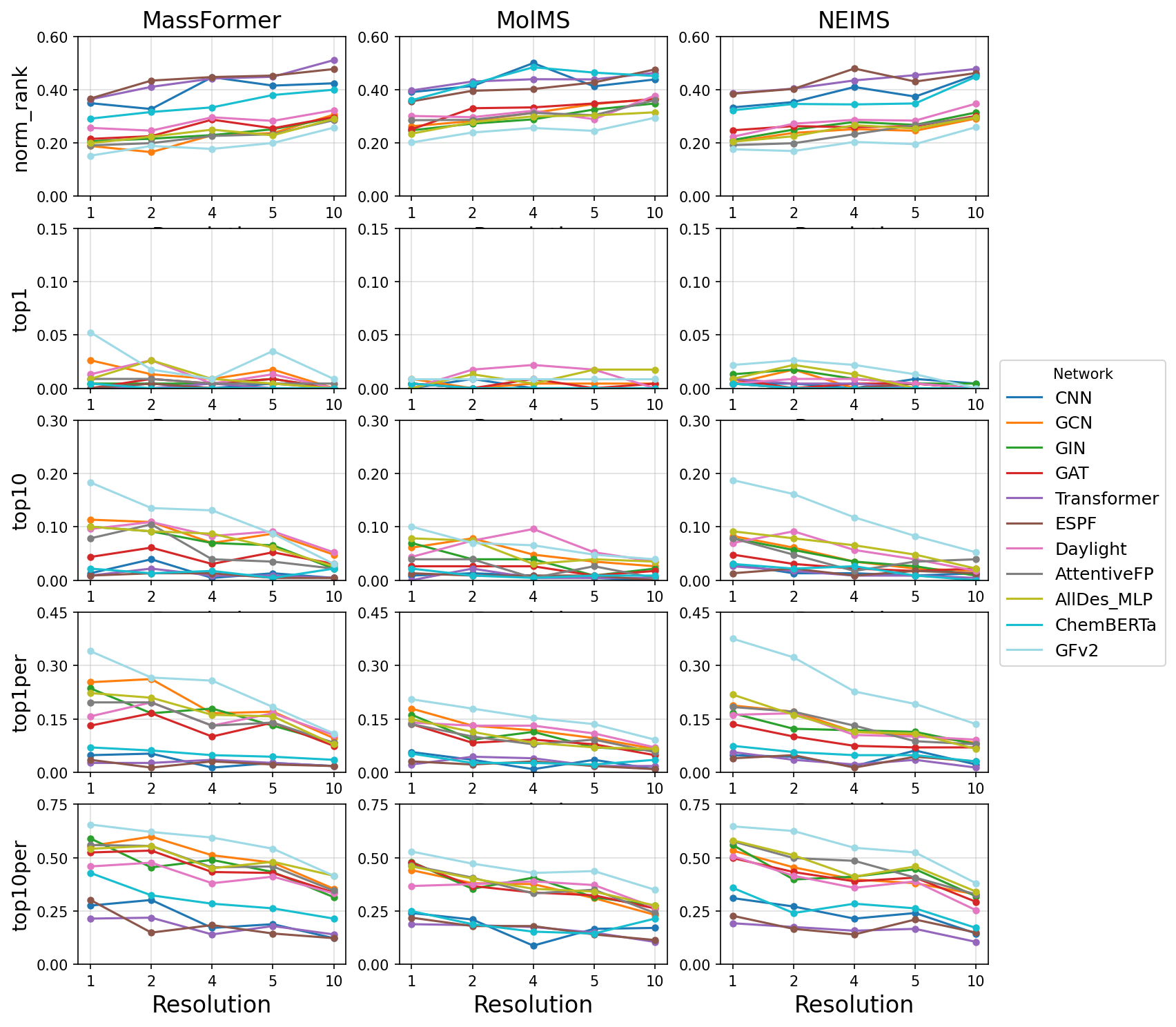}
    \caption{Performance of different embedder-predictor combinations with different resolutions on the CASMI 2022 retrieval benchmark.} 
\label{fig:retrieval-casmi2022}
\end{figure}

\section{Future Directions}
\label{sec:future_direction}
Several directions remain open for extending the benchmark scope of FlexMS. A first priority is to move beyond the current binned formulation toward evaluation settings that better reflect the structured nature of tandem mass spectra, including peak-list prediction, formula-aware decoding, and other variable-length output formulations. Such extensions would test whether conclusions that hold for binned regression remain stable under more chemically explicit prediction targets.

A second direction concerns benchmark coverage under domain shift. Although the present study examines transfer across ion types and instrument classes, important sources of variation remain underexplored, including collision-energy differences and other conditions. More systematic protocols for these cases would strengthen the benchmark as a testbed for robust generalization rather than only within-protocol comparison.

A third direction is to broaden efficiency-aware evaluation. Current benchmark tables emphasize predictive quality, but practical adoption also depends on training cost, inference latency, and memory footprint. Standardized budget-aware reporting would make it easier to compare architectures under realistic deployment constraints and to distinguish accuracy gains from gains that remain usable at scale.

Finally, future benchmark iterations would benefit from wider dataset coverage and more diverse downstream tasks. Public MS/MS resources remain uneven in chemical space, adduct composition, acquisition settings, and annotation quality. As additional curated resources become available, the benchmark could be extended to cover broader molecule classes and more realistic identification to do reproducible comparison.

\clearpage

\end{document}